\documentclass[twoside,twocolumn,9pt]{article}
\usepackage{extsizes}
\usepackage[super,sort&compress,comma]{natbib} 
\usepackage[version=3]{mhchem}
\usepackage[left=1.5cm, right=1.5cm, top=1.785cm, bottom=2.0cm]{geometry}
\usepackage{balance}
\usepackage{mathptmx}
\usepackage{sectsty}
\usepackage{graphicx} 
\usepackage{lastpage}
\usepackage[format=plain,justification=justified,singlelinecheck=false,font={stretch=1.125,small,sf},labelfont=bf,labelsep=space]{caption}
\usepackage{float}
\usepackage{fancyhdr}
\usepackage{fnpos}
\usepackage[english]{babel}
\addto{\captionsenglish}{%
  
}
\usepackage{array}
\usepackage{droidsans}
\usepackage{charter}
\usepackage[T1]{fontenc}
\usepackage{subcaption}
\usepackage[usenames,dvipsnames]{xcolor}
\usepackage{setspace}
\usepackage[compact]{titlesec}
\usepackage{hyperref}

\usepackage{epstopdf}
\usepackage{booktabs}
\usepackage{multirow}

\definecolor{cream}{RGB}{222,217,201}

\begin{document}

\pagestyle{fancy}
\thispagestyle{plain}
\fancypagestyle{plain}{
\renewcommand{\headrulewidth}{0pt}
}

\makeFNbottom
\makeatletter
\renewcommand\LARGE{\@setfontsize\LARGE{15pt}{17}}
\renewcommand\Large{\@setfontsize\Large{12pt}{14}}
\renewcommand\large{\@setfontsize\large{10pt}{12}}
\renewcommand\footnotesize{\@setfontsize\footnotesize{7pt}{10}}
\makeatother

\renewcommand{\thefootnote}{\fnsymbol{footnote}}
\renewcommand\footnoterule{\vspace*{1pt}%
\color{cream}\hrule width 3.5in height 0.4pt \color{black}\vspace*{5pt}} 
\setcounter{secnumdepth}{5}

\makeatletter 
\renewcommand\@biblabel[1]{#1}            
\renewcommand\@makefntext[1]%
{\noindent\makebox[0pt][r]{\@thefnmark\,}#1}
\makeatother 
\renewcommand{\figurename}{\small{Fig.}~}
\sectionfont{\sffamily\Large}
\subsectionfont{\normalsize}
\subsubsectionfont{\bf}
\setstretch{1.125} 
\setlength{\skip\footins}{0.8cm}
\setlength{\footnotesep}{0.25cm}
\setlength{\jot}{10pt}
\titlespacing*{\section}{0pt}{4pt}{4pt}
\titlespacing*{\subsection}{0pt}{15pt}{1pt}

\fancyfoot{}
\fancyfoot[LO,RE]{\vspace{-7.1pt}\includegraphics[height=9pt]{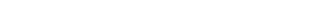}}
\fancyfoot[CO]{\vspace{-7.1pt}\hspace{13.2cm}\includegraphics{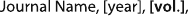}}
\fancyfoot[CE]{\vspace{-7.2pt}\hspace{-14.2cm}\includegraphics{head_foot/RF}}
\fancyfoot[RO]{\footnotesize{\sffamily{1--\pageref{LastPage} ~\textbar  \hspace{2pt}\thepage}}}
\fancyfoot[LE]{\footnotesize{\sffamily{\thepage~\textbar\hspace{3.45cm} 1--\pageref{LastPage}}}}
\fancyhead{}
\renewcommand{\headrulewidth}{0pt} 
\renewcommand{\footrulewidth}{0pt}
\setlength{\arrayrulewidth}{1pt}
\setlength{\columnsep}{6.5mm}
\setlength\bibsep{1pt}

\makeatletter 
\newlength{\figrulesep} 
\setlength{\figrulesep}{0.5\textfloatsep} 

\newcommand{\topfigrule}{\vspace*{-1pt}%
\noindent{\color{cream}\rule[-\figrulesep]{\columnwidth}{1.5pt}} }

\newcommand{\botfigrule}{\vspace*{-2pt}%
\noindent{\color{cream}\rule[\figrulesep]{\columnwidth}{1.5pt}} }

\newcommand{\dblfigrule}{\vspace*{-1pt}%
\noindent{\color{cream}\rule[-\figrulesep]{\textwidth}{1.5pt}} }

\makeatother

\twocolumn[
  \begin{@twocolumnfalse}
{\includegraphics[height=30pt]{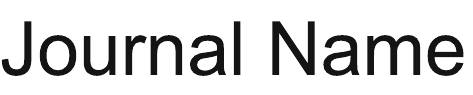}\hfill\raisebox{0pt}[0pt][0pt]{\includegraphics[height=55pt]{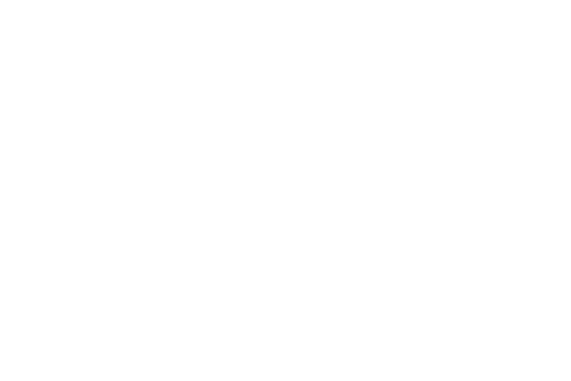}}\\[1ex]
\includegraphics[width=18.5cm]{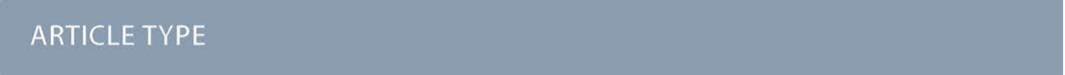}}\par
\vspace{1em}
\sffamily
\begin{tabular}{m{4.5cm} p{13.5cm} }

\includegraphics{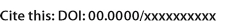} & \noindent\LARGE{\textbf{Quantification and Classification of Carbon Nanotubes in Electron Micrographs using Vision Foundation Models$^\dag$}} \\
\vspace{0.3cm} & \vspace{0.3cm} \\

 & \noindent\large{Sanjay Pradeep\textit{$^{a}$}, Chen Wang\textit{$^{b}$}, Matthew M. Dahm\textit{$^{c}$}, Jeff D. Eldredge$^{\ast}$\textit{$^{a}$}, and Candace S. J. Tsai$^{\ast}$\textit{$^{d}$}} \\

\includegraphics{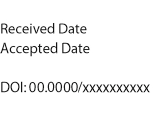} & \noindent\normalsize{Accurate characterization of carbon nanotube morphologies in electron microscopy images is vital for exposure assessment and toxicological studies, yet current workflows rely on slow, subjective manual segmentation. This work presents a unified framework leveraging vision foundation models to automate the quantification and classification of CNTs in electron microscopy images. First, we introduce an interactive quantification tool built on the Segment Anything Model (SAM) that segments particles with near-perfect accuracy using minimal user input. Second, we propose a novel classification pipeline that utilizes these segmentation masks to spatially constrain a DINOv2 vision transformer, extracting features exclusively from particle regions while suppressing background noise. Evaluated on a dataset of 1,800 TEM images, this architecture achieves 95.5\% accuracy in distinguishing between four different CNT morphologies, significantly outperforming the current baseline despite using a fraction of the training data. Crucially, this instance-level processing allows the framework to resolve mixed samples, correctly classifying distinct particle types co-existing within a single field of view. These results demonstrate that integrating zero-shot segmentation with self-supervised feature learning enables high-throughput, reproducible nanomaterial analysis, transforming a labor-intensive bottleneck into a scalable, data-driven process.} \\

\end{tabular}

\end{@twocolumnfalse} \vspace{0.6cm}]


\renewcommand*\rmdefault{bch}\normalfont\upshape
\rmfamily
\section*{}
\vspace{-1cm}

\footnotetext{\textit{$^{a}$~Mechanical and Aerospace Engineering, University of California, Los Angeles, Los Angeles, CA 90095, USA. E-mail: jdeldre@ucla.edu}}
\footnotetext{\textit{$^{b}$~Health Effects Laboratory Division, National Institute for Occupational Safety and Health, 1090 Tusculum Ave, Cincinnati, OH 45226, USA.}}
\footnotetext{\textit{$^{c}$~Office of the Director, National Institute for Occupational Safety and Health, 1090 Tusculum Ave, Cincinnati, OH 45226, USA.}}
\footnotetext{\textit{$^{d}$~Fielding School of Public Health, University of California, Los Angeles, Los Angeles, CA 90095, USA.  E-mail: candacetsai@ucla.edu}}
\footnotetext{\dag~Electronic Supplementary Information (ESI) available: Detailed architecture specifications, additional results, and supplementary figures. See DOI: 00.0000/00000000.}

\section{Introduction}

Carbon nanotubes (CNTs) and carbon nanofibers (CNFs) are a rapidly expanding class of engineered carbon structures whose exceptional mechanical, electrical, and thermal properties are being harnessed across diverse technologies\cite{SHVEDOVA2009192}. From carbon-fiber bicycle frames and sporting goods to advanced electronics, aerospace structures, and energy devices, CNTs and related carbon nanomaterials (CNMs) are increasingly embedded in products across both consumer and high-technology sectors. As these materials transition from laboratory-scale research to large-volume industrial production and widespread use, the potential for human and environmental exposure increases accordingly. This is a significant concern, as toxicological studies have indicated that certain types of CNTs, particularly those with high aspect ratios, can elicit harmful biological responses upon inhalation, including persistent pulmonary inflammation and fibrosis, drawing parallels to asbestos-like pathogenicity \cite{KatherineDuke, DONALDSON20132078}. Unlike many other engineered nanomaterials, carbon nanomaterials can differ greatly in their physical and chemical properties depending on how they are synthesized and treated after production, and recent epidemiological and toxicological studies suggest that fiber dimensions and the surface properties of agglomerates or bundles are important determinants of their toxic effects \cite{Schulte2019-jv,NIOSH_CIB65_2013, Birch_CNT_TEM_2017, Oberdorster_CNTCNF_2015}. These findings underscore the critical need for improved particle identification, counting, and classification protocols to enable meaningful exposure and risk assessment. Accurate morphological characterization of airborne particles is essential for occupational health surveillance, particularly for quantifying the respirable fraction and evaluating the biological hazard potential of CNMs in workplace environments where exposure may occur.

Scanning Electron Microscopy (SEM) and Transmission Electron Microscopy (TEM) are the typical visualization methods for characterizing micron to nanometer sized carbon nanomaterials, particularly in occupational exposure studies where particles are collected on sampling media such as filters, mesh or TEM grids according to established sampling methods or NIOSH protocols\cite{Birch_CNT_TEM_2017}. While these techniques yield images rich in quantitative and qualitative data, extracting actionable insights presents a significant and well-documented bottleneck. The analysis---comprising particle identification, quantification, and morphological classification---traditionally relies on manual or semi-automated workflows using tools like ImageJ \cite{ImageJ} and Fiji\cite{Fiji}. To isolate particles from porous filters or complex backgrounds, researchers must often develop complex, multi-step image processing workflows involving advanced thresholding, morphological operations and watershedding. These workflows are notoriously brittle, often failing when applied to images with varying illumination, background textures, or particle agglomeration states. Consequently, this process is not only exceptionally time-consuming but is also subject to significant inter-operator variability, limiting the throughput and statistical reliability required for robust risk assessment.

To overcome the limitations of manual analysis, the field has increasingly turned to machine learning (ML) and computer vision. Early efforts demonstrated the potential of Convolutional Neural Networks (CNNs) for these tasks\cite{deep_microscopy,deep_microscopy2,deep_microscopy3,deep_microscopy4,Xie_Hopke_Wienke_1994,Wienke_3MAP_1995,Trujillo2017}. For instance, deep learning based detection and segmentation models such as YOLO \cite{yolo} and U-Net\cite{unet} have been adapted for particle detection and pixel-wise segmentation in SEM and TEM images\cite{yolo_TEM,unet_EM}, but often struggle with the dense, overlapping, and morphologically diverse nature of nanoparticle agglomerates, and typically require extensive, manually-annotated bounding boxes for training\cite{yolo_drawback1,yolo_drawback2}. For the more complex task of classification, Luo et al. (2021) \cite{Luo_Wang} introduced a transfer learning approach that combined a VGG-16 architecture \cite{vgg} with a Vector of Locally Aggregated Descriptors (VLAD) \cite{arandjelovic2016netvlad} to classify TEM images of CNTs. While a significant step toward automation, this method highlighted a key challenge: its feature extraction, which sampled random pixels from the entire image, required a large dataset of over 5300 images and extensive data augmentation to achieve high accuracy, and cannot be applied to images with multiple types of CNT. The high data dependency of these models has remained a significant barrier to the accessible deployment of deep learning models in materials science and occupational exposure labs.

The recent advent of large-scale, pre-trained foundation models for computer vision, however, represents a paradigm shift. Models like the Segment Anything Model (SAM) \cite{kirillov2023segment} are trained on millions of images with billions of segmentation masks, enabling them to perform "zero-shot" or "few-shot" segmentation of any object in an image with remarkable precision. This technology has already been recognized as a powerful tool for scientific imaging. Archit et al. (2025) \cite{Archit2025} demonstrated this by developing "Segment Anything for Microscopy (µSAM)," a fine-tuned version of SAM for light and electron microscopy that can be used as an interactive tool, effectively replacing the laborious manual segmentation of traditional software. Concurrently, self-supervised models like DINOv2 \cite{dinov2} have learned to produce highly discriminative feature representations from images without human labels. This combination of segmentation and feature extraction using foundation models has the potential to be a state-of-the-art approach. For example, Barnatan et al. (2025) \cite{Bartanan} recently proposed a zero-shot shape classification pipeline for nanoparticles by combining SAM-generated segments with DINOv2 feature embeddings, further validating the power of this coupled approach.

\begin{figure}[t!]
    \centering
    \includegraphics[width=\linewidth]{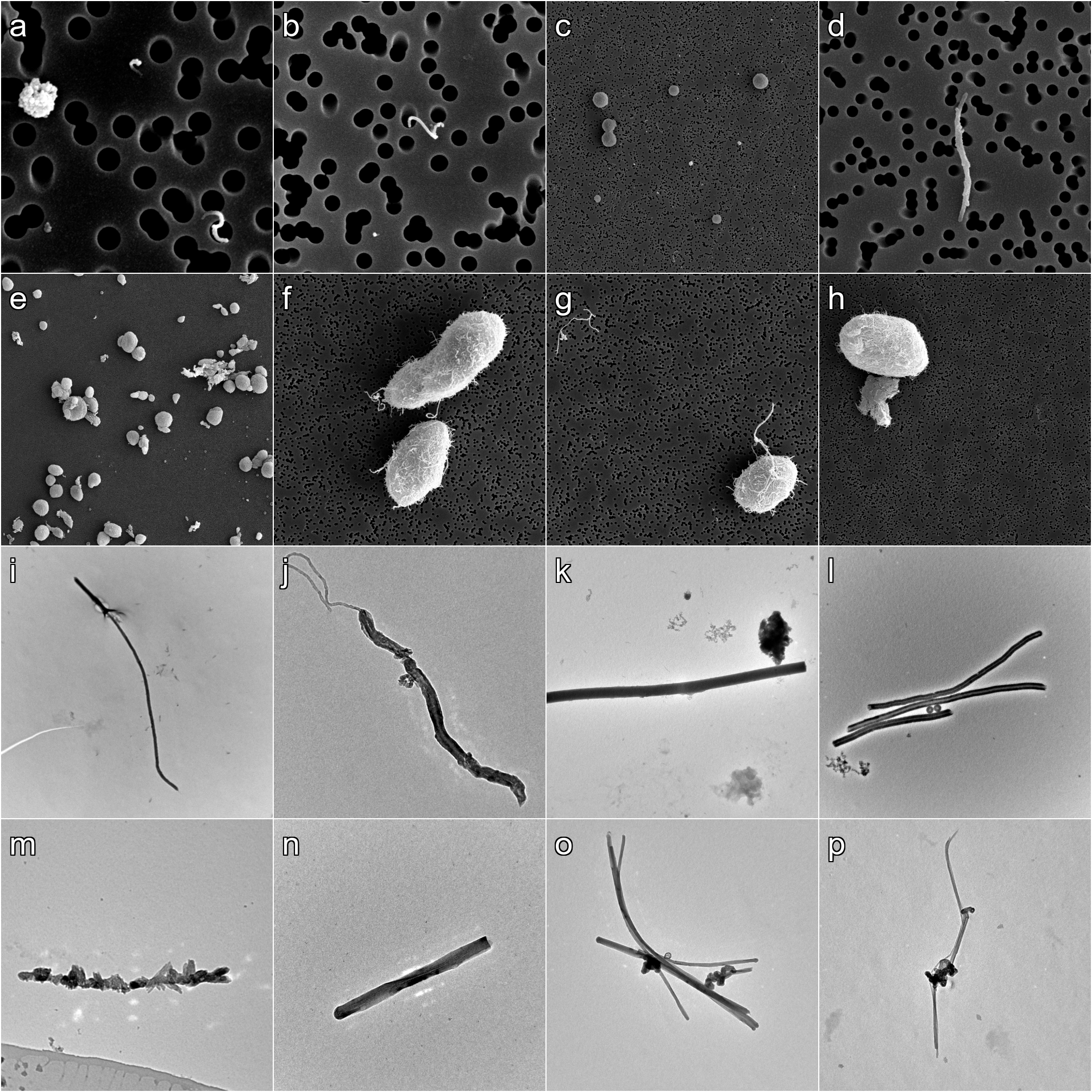}
    \caption{Representative microscopy images from the segmentation validation dataset. (a--h) SEM images of hydrophilic and hydrophobic CNT samples on polycarbonate filters displaying diverse aggregation states and morphologies including isolated particles, small clusters, and large agglomerates. (i--p) TEM images showing individual CNT structures with varying morphologies including single fibers, bundles and clusters.}
\label{fig:seg_dataset}    
\end{figure}

\begin{figure}[t!]
    \centering
    \includegraphics[width=\linewidth]{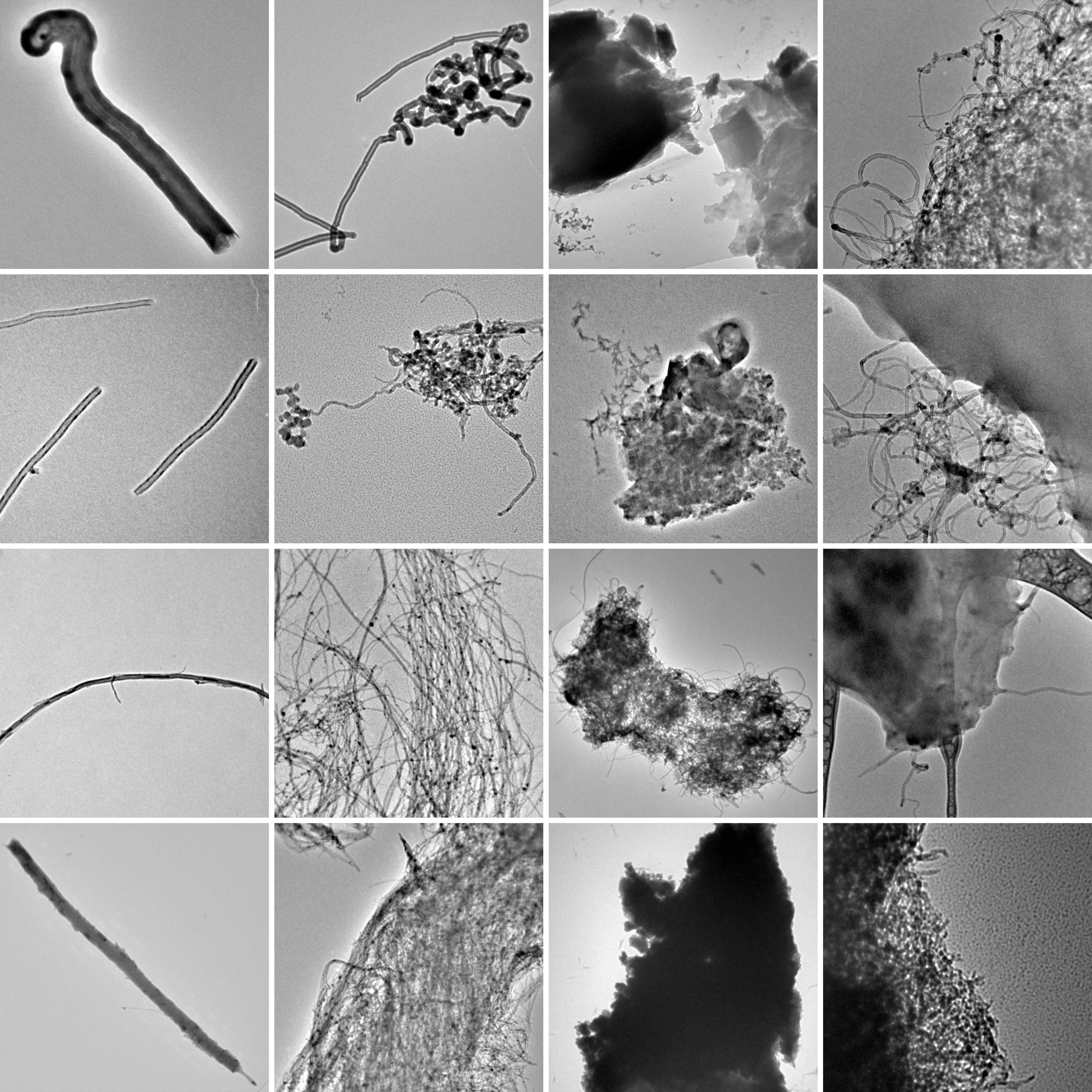}
    
    \vspace{0.5mm} 
    
    \makebox[0.25\linewidth]{(a)}\hfill%
    \makebox[0.25\linewidth]{(b)}\hfill%
    \makebox[0.25\linewidth]{(c)}\hfill%
    \makebox[0.25\linewidth]{(d)}
    
    \caption{Classification dataset (NIOSH): Representative TEM images organized by morphology class. (a) Fiber: elongated structures with high aspect ratios, (b) Cluster: dense, non-linear agglomerates with entangled CNTs, (c) Matrix: particles embedded within web-like matrices, and (d) Matrix Surface: CNTs extruding from particle surfaces.}
    \label{fig:class_dataset}
\end{figure}

In this work, we harness these synergistic foundation models to build a comprehensive framework for both the quantification and classification of carbon nanotubes from electron micrographs. Our approach is a multi-stage process designed to first enable rapid, high-quality segmentation and quantification, and then to leverage these segments for robust, data-efficient classification. Particle segmentation is already a necessary step in the quantification workflow for counting and measuring individual CNT structures; our methodology integrates automated morphological classification directly into this existing process, allowing researchers to obtain both quantitative metrics and categorical labels simultaneously without additional manual effort. We present and validate a semi-automated, SAM-based interactive tool for the rapid and accurate segmentation of CNTs from diverse SEM and TEM images as shown in Fig~\ref{fig:seg_dataset}. This tool is designed to replace traditional manual methods, allowing researchers with no computational background to perform robust particle counting and sizing. We then introduce a novel, data-efficient classification pipeline that leverages DINOv2 (ViT-B/14) for feature extraction. By guiding the DINOv2 encoder to extract features only from the segmented particles, we eliminate background noise and enable a simple MLP classifier to achieve state-of-the-art performance in categorizing particles into classes such as fiber, cluster, matrix, and matrix surface as shown in Fig~\ref{fig:class_dataset}. We demonstrate that this segmentation-guided pipeline achieves 95.5\% classification accuracy while using only a fraction of the training data required by previous benchmarks~\cite{Luo_Wang}, offering a generalizable and accessible framework for the rapid and reliable quantification and characterization of CNTs.

The remainder of this paper is organized as follows: Section 2 describes the datasets used for segmentation validation and classification training, followed by detailed descriptions of the SAM-based segmentation pipeline and the DINOv2 classification architecture. Section 3 presents segmentation performance results across different SAM model variants and imaging modalities, systematic classification experiments comparing 24 model configurations, and visualization analyses explaining the learned representations. Section 4 concludes with a discussion of the framework's broader applicability to nanomaterial characterization and future directions for domain-specialized foundation models.

\section{Dataset and Methods}
This work uses two distinct, purpose-built datasets designed to validate our segmentation framework and train our classification pipeline. For assessing the segmentation performance of the interactive SAM tool, we curated an in-house collection of 200 electron microscope images evenly distributed between Scanning Electron Microscopy (SEM) and Transmission Electron Microscopy (TEM) modalities. This segmentation validation dataset deliberately captures the full spectrum of imaging conditions and sample characteristics encountered in CNT exposure assessment \cite{Birch_CNT_TEM_2017}, including samples collected on polycarbonate and mixed cellulose filters from open-face and closed-face cassettes \cite{NIOSH_NMAM_AE_2016, Beaulieu1980OpenClosed}, and Tsai Diffusion Samplers \cite{tsai2018sampler}. The SEM images contain both industry-grade multi-walled CNTs (MWCNT) with outer diameter 15 $\pm$ 5 nm, length 5--20 $\mu$m and hydroxyl-functionalized hydrophilic MWCNTs with outer diameter 10--20 nm and length 10--30 $\mu$m. The images exhibit substantial morphological diversity from isolated fibers to complex clusters and dense agglomerates, alongside considerable variation in substrate texture and background contrast, as seen in Fig. \ref{fig:seg_dataset}, both of which commonly challenge conventional segmentation algorithms. For each image, we generated detailed ground-truth segmentation masks through a combination of Fiji-based segmentation and manual refinement to precisely delineate particle boundaries.

For the classification task, we employed a subset of 1,800 images from a comprehensive repository of 5,300 grayscale TEM images of various CNT/CNF materials from 26 facilities collected by the U.S. National Institute for Occupational Safety and Health (NIOSH) \cite{HarvardDataverse,Dahm_CNT_exposure_2015,Dahm2018-vh}. By utilizing this dataset, our study targets the classification of complex industrial field samples, as opposed to lab-generated samples with distinct morphological features. We generated segmentation masks for each selected image using the interactive SAM tool and assigned expert-annotated labels corresponding to one of four morphologically distinct classes that reflect the native structures of CNTs/CNFs and alternative structural forms commonly observed in field samples. A "fiber" represents cylindrical carbon nanostructures with various stacking arrangements of graphene sheets and aspect ratios exceeding 3:1, consistent with NIOSH counting criteria for airborne fibers associated with elevated toxicological potential \cite{niosh7400}. A "cluster" denotes bundles of such fibers that are loosely contacted, formed through inter-fiber interactions and the inherent flexibility of CNT/CNFs, which are known to agglomerate easily. A "matrix" describes the most common morphology in workplace air samples, where CNT/CNFs agglomerate and pack into condensed, web-like structures that obscure individual particle boundaries. Finally, a "matrix surface" identifies CNT/CNF fibers embedded within large particles comprising other carbonaceous materials, catalytic particles, and intermediate products emitted during manufacturing operations, where tube/fiber structures extrude from the particle surface and remain discernible at higher magnifications\cite{Birch_CNT_TEM_2017}.

The classification dataset comprises 1,800 labeled TEM images with perfect class balance, 450 images each of Cluster, Fiber, Matrix, and Matrix Surface morphologies, ensuring unbiased model training across all categories. We partitioned the dataset into training (80\%, 1,440 images), validation (10\%, 180 images), and test (10\%, 180 images) sets using stratified sampling to preserve class proportions across splits. This widely adopted train/validation/test split enables systematic comparison of different classification architectures under identical experimental conditions.

\subsection{Segmentation Pipeline Architecture}
\label{sec:pipeline_overview}

\begin{figure*}[htbp]
    \centering
    \includegraphics[width=\textwidth]{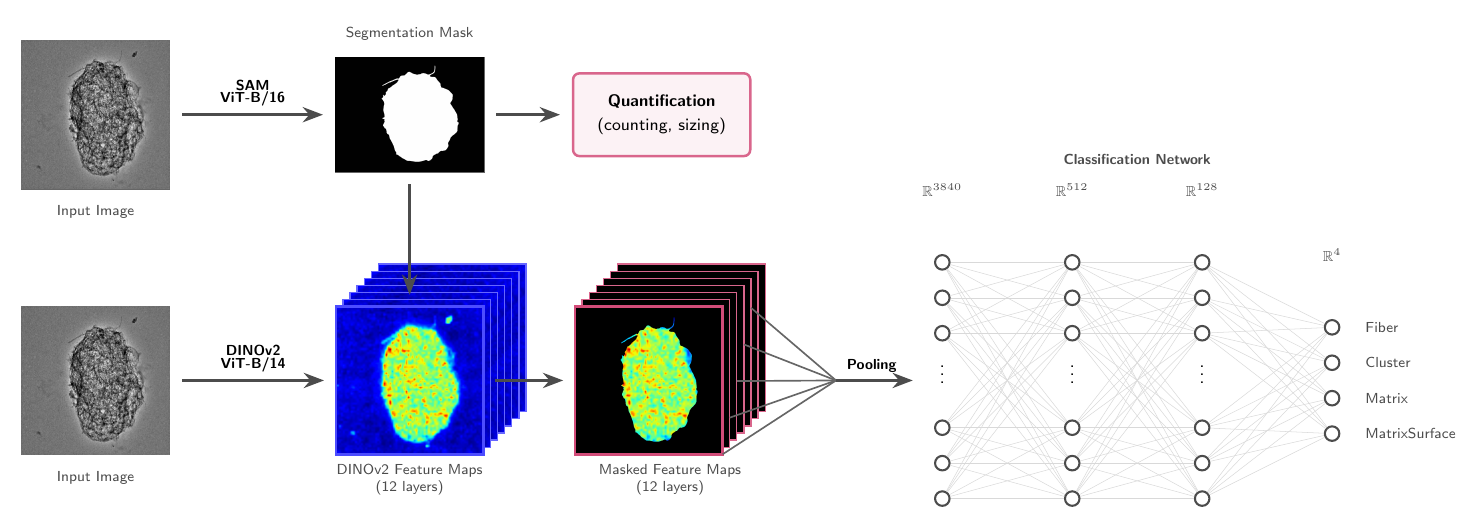}
    \caption{Dual-pathway architecture for CNT morphology analysis integrating SAM and DINOv2. The Segment Anything Model (SAM ViT-B/16) generates binary segmentation masks from input SEM images for downstream quantification tasks (counting, sizing). DINOv2 (ViT-B/14) extracts multi-scale feature representations across 12 transformer blocks, visualized as stacked activation maps. The segmentation mask is applied to suppress background regions, creating masked feature maps (red borders) that focus on CNT morphology. Features from multiple layers are aggregated through hypercolumn pooling into a 3840-dimensional vector, then classified via a fully-connected network to predict four morphology classes: Fiber, Cluster, Matrix, or MatrixSurface. The convergent arrows illustrate multi-scale feature pooling from different network depths. \textit{Note:} Input image dimensions differ between pathways due to distinct patch size requirements (SAM: 16$\times$16, DINOv2: 14$\times$14).}
    \label{fig:architecture}
\end{figure*}

In the first step, the SAM-based interactive tool implements a streamlined click-based workflow where users load electron microscope images. When scale bars are present (as is standard in exposure assessment studies), the system automatically extracts calibration information using easyOCR \cite{easyocr} text detection, eliminating manual scale entry. Upon loading, the image is passed through the SAM ViT (Vision Transformer) encoder to generate a high-dimensional feature embedding, a computationally intensive operation performed only once per image and cached for all subsequent interactions. Users then interact with the image through intuitive prompts: positive clicks (left-click) on target particles and optional negative clicks (right-click) to exclude undesired regions. With each prompt, the lightweight decoder operates on the pre-computed embedding to generate segmentation masks in milliseconds, enabling real-time iterative refinement until precise particle boundaries are achieved. The resulting masks undergo automated post-processing using scikit-image \cite{van2014scikit} morphological operations including \texttt{remove\_small\_objects()} for noise filtering and custom functions to eliminate edge artifacts. Connected component analysis via \texttt{measure.label()} identifies individual particles, and \texttt{regionprops()} extracts comprehensive morphological descriptors for each segmented object.  Our approach leverages SAM's powerful zero-shot capabilities within a human-in-the-loop workflow, removing the need for per-image parameter tuning while remaining accessible to users without scientific image analysis expertise.

To validate segmentation performance and efficiency, we conducted a formal experiment using our 200-image validation dataset spanning diverse SEM and TEM imaging conditions. A user segmented the primary particle(s) in each image from scratch, and resulting masks were compared against manually created ground truth using standard metrics including Dice coefficient, Intersection-over-Union (IoU), precision, recall, and accuracy. We also recorded the total number of clicks required per image as a quantitative measure of user effort, demonstrating the minimal interaction needed compared to traditional workflows that often require several minutes per image.

Following segmentation and post-processing, the tool automatically calculates morphological parameters essential for exposure assessment and materials characterization. Using the extracted scale calibration (nm/pixel), the system computes particle count from labeled connected components, individual particle areas, equivalent diameters, aspect ratios, and Feret diameters. This automated pipeline, from scale extraction through quantification, directly replaces time-consuming manual measurements, significantly improving throughput and reproducibility while eliminating subjective variability inherent in traditional workflows.

\subsection{Classification Pipeline Architecture}
\label{sec:methods_classification}

Figure~\ref{fig:architecture} illustrates our dual-pathway architecture that integrates SAM-based segmentation with DINOv2 feature extraction for CNT morphology classification. The second component of our methodology is a classification pipeline that leverages the segmentation masks from SAM to guide feature extraction. Unlike conventional approaches that extract features uniformly across entire images, including task-irrelevant background regions like filter substrates, our segmentation-guided strategy samples features only from pixels corresponding to CNT structures of interest. Ground-truth masks are downsampled to the feature map resolution of each encoder, and we sample pixels within masked regions to obtain variable-length sets of local feature descriptors. This focused sampling suppresses background noise, increases the density of informative features, and allows effective learning from comparatively small training datasets by reducing the need for the model to learn invariance to background variations. We apply this mask-guided sampling consistently across all architectures, and in parallel we implement a baseline approach that performs uniform sampling across entire images (without using segmentation masks), thereby providing a direct and controlled comparison between mask-guided and uniform feature extraction strategies.

We evaluate two complementary vision transformer encoders for feature extraction: SAM ViT-B/16 and DINOv2 ViT-B/14, which differ fundamentally in their patch tokenization strategies. SAM's ViT-B architecture uses 16×16 pixel patches to divide input images into discrete tokens, while DINOv2's ViT-B/14 employs smaller 14×14 patches. Rather than using only the final layer representations, we extract multi-scale "hypercolumn" features from five intermediate transformer blocks (layers 1, 3, 6, 9, and 11 of 12 total layers), capturing a hierarchical range of visual information from low-level texture patterns to high-level semantic structures. Each transformer block outputs 768-dimensional feature vectors at every spatial location; concatenating features from all five layers yields a 3,840-dimensional hypercolumn descriptor per pixel. In our implementation, we processed images at each encoder's native resolution: SAM at 1024×1024(producing 64×64 feature maps) and DINOv2 at 518×518 (producing 37×37 feature maps). For mask-guided sampling, we downsampled the binary segmentation masks to match each encoder's respective feature map resolution, then sampled hypercolumn descriptors only from spatial locations corresponding to CNT particles. This approach allows us to systematically compare how features learned for general-purpose segmentation (SAM) versus self-supervised visual representation (DINOv2) perform when adapted to nanomaterial classification through our mask-guided sampling strategy.

To convert these variable-length sets of sampled features into fixed-length representations suitable for classification, we systematically evaluate three pooling strategies that aggregate the hypercolumn descriptors in different ways. The first is average pooling, which computes the mean of all sampled hypercolumn descriptors, yielding a single 3,840-dimensional embedding that captures the typical appearance of CNT structures. The second is max pooling, which takes the maximum activation across sampled descriptors for each feature dimension, emphasizing the most prominent patterns present in the particle. The third is a concatenation of both average and max pooling (avg+max), producing a 7,680-dimensional representation that combines information about both typical and extreme feature activations within the CNT region. Each pooled representation is standardized to zero mean and unit variance before being passed to a classifier. These three pooling strategies are applied identically to both SAM and DINOv2 features, and are evaluated under both mask-guided and uniform sampling conditions, enabling systematic comparison of aggregation methods across different feature extraction approaches.

On top of these pooled embeddings, we consider two classifier families that differ in complexity but share the same input representation. The first is a linear classifier (multinomial logistic regression) operating directly on the pooled embedding (3,840 dimensions for single pooling, 7,680 for avg+max), representing a standard ``linear probe'' on top of frozen foundation-model features. This simple classifier serves as a baseline that measures how linearly separable the pooled features are. The second is a shallow 2-layer Multi-Layer Perceptron (MLP) with architecture $d \to 512 \to 128 \to 4$ (where $d$ is the input dimension), incorporating batch normalization and ReLU activations in the hidden layers, and 0.3 dropout for regularization. The MLP is trained end-to-end on the frozen embeddings using the Adam optimizer, mini-batch stochastic gradient descent, and early stopping based on validation loss to prevent overfitting. Cross-entropy loss is used for training, with the output layer producing logits that are converted to class probabilities via the softmax function. These two classifier heads allow us to quantify how much performance gain comes from the representation quality (determined by the encoder, sampling strategy, and pooling method) versus the capacity of the classifier itself (linear versus shallow non-linear).

Combining the two encoders (SAM, DINOv2), two sampling strategies (mask-guided, uniform), three pooling methods (avg, max, avg+max), and two classifier types (linear, MLP) yields a comprehensive experimental design with 24 distinct model configurations. For each encoder, we train six models under mask-guided sampling (three pooling strategies $\times$ two classifiers) and six models under uniform sampling (three pooling strategies $\times$ two classifiers), enabling systematic analysis of each design choice. The mask-guided models form the core of our study, as they directly leverage the segmentation information to focus feature extraction on CNT structures. The uniform sampling models serve as controlled baselines that quantify the benefit of segmentation guidance by processing the same images with the same architectures but without using mask information. This experimental design allows us to isolate and measure the individual contributions of (i) foundation model choice (SAM versus DINOv2), (ii) the impact of segmentation-guided versus uniform feature sampling, (iii) pooling strategy (average, max, or their combination), and (iv) classifier complexity (linear versus shallow MLP).

All models are trained and evaluated on identical stratified 80/10/10 train/validation/test splits with fixed random seeds for reproducibility. The same data splits are used across all model configurations to ensure fair comparison. The validation set is used exclusively for hyperparameter tuning during development. Once optimal hyperparameters are identified, the final model for each configuration is retrained on the combined training and validation set (1,620 images) and evaluated on the held-out test set (180 images) to obtain final performance metrics. We report accuracy, macro-averaged F1 score, and present confusion matrices to analyze per-class performance.

\begin{figure}[b!]
    \centering
    \includegraphics[width=\linewidth]{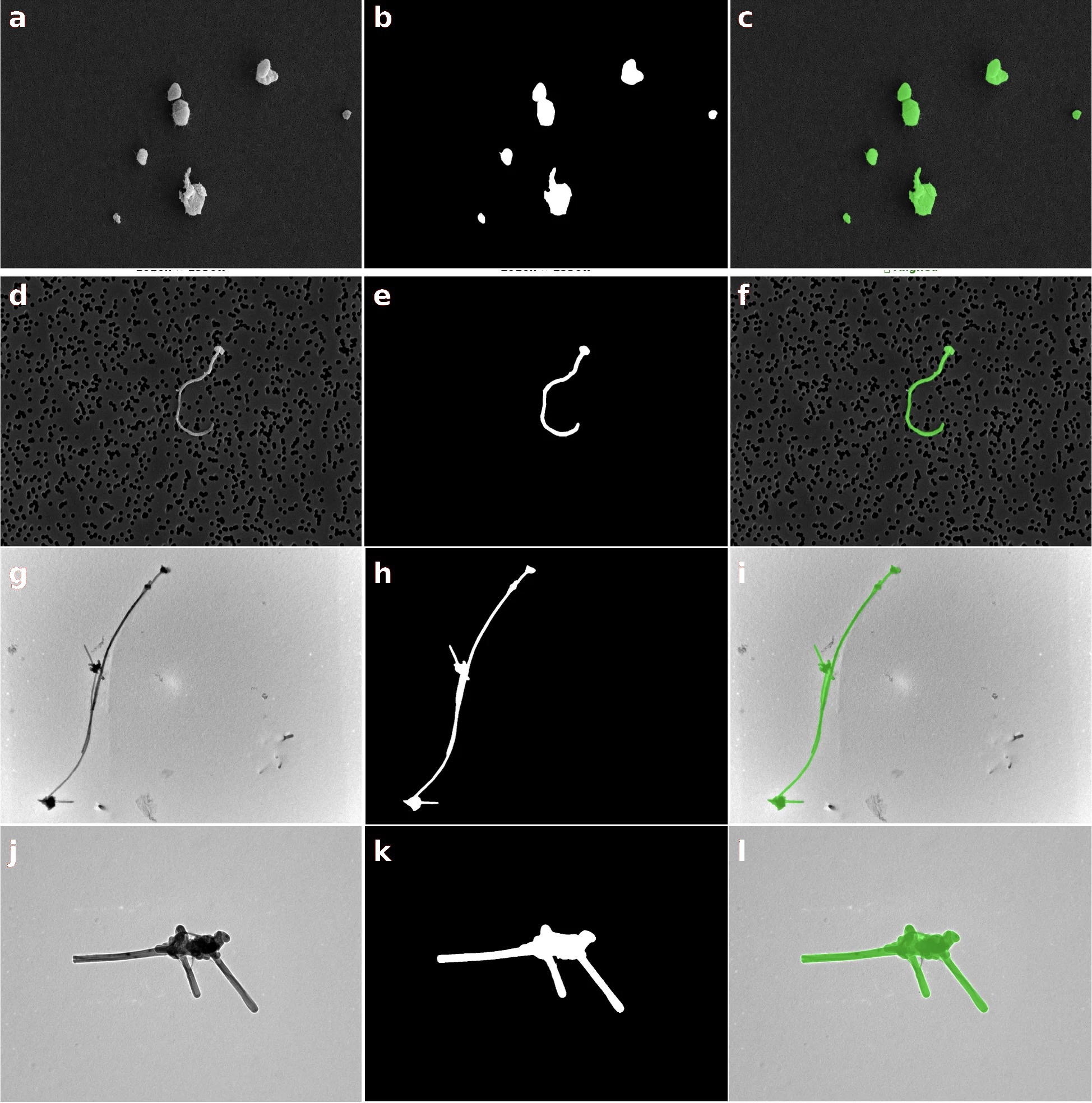}
    \caption{\small\sffamily Comparison of original microscopy images, ground truth segmentation masks, and overlay visualizations for different sample types. For each row, the left panel shows the original microscopy image (SEM for rows 1-2, TEM for rows 3-4), the center panel presents the binary ground truth mask, and the right panel displays the ground truth overlay in green on the original image, highlighting the segmented regions of interest.}
    \label{fig:microscopy_comparison}
    
\end{figure}

\section{Results and Discussion}

\subsection{Segmentation Performance Across Model Architectures and Imaging Modalities}

\begin{table}[t!]
\centering
\small
\setlength{\tabcolsep}{4pt}
\renewcommand{\arraystretch}{1.5}  
\caption{\small\sffamily Segmentation performance metrics and user effort comparison across SAM model architectures (ViT-H and ViT-B) and electron microscopy imaging modalities (SEM and TEM).}
\label{tab:segmentation_results}
\begin{tabular}{lcccc}
\hline
\textbf{Metric} & \textbf{SEM ViT-H} & \textbf{SEM ViT-B} & \textbf{TEM ViT-H} & \textbf{TEM ViT-B} \\
\hline
Dice & $0.977 \pm 0.026$ & $0.954 \pm 0.054$ & $0.982 \pm 0.021$ & $0.970 \pm 0.023$ \\
IoU  & $0.957 \pm 0.048$ & $0.917 \pm 0.087$ & $0.965 \pm 0.039$ & $0.943 \pm 0.042$ \\
Precision & $0.983 \pm 0.033$ & $0.961 \pm 0.068$ & $0.985 \pm 0.019$ & $0.969 \pm 0.032$ \\
Recall & $0.973 \pm 0.032$ & $0.951 \pm 0.056$ & $0.979 \pm 0.037$ & $0.972 \pm 0.029$ \\
Accuracy & $0.999 \pm 0.002$ & $0.998 \pm 0.003$ & $0.999 \pm 0.002$ & $0.998 \pm 0.002$ \\
\hline
Clicks & $6.37 \pm 4.66$ & $5.53 \pm 3.40$ & $3.66 \pm 1.94$ & $4.77 \pm 2.20$ \\
\end{tabular}
\end{table}

The Segment Anything Model (SAM) is available in three backbone architectures of increasing complexity: ViT-Base (ViT-B, 91M parameters), ViT-Large (ViT-L, 308M parameters), and ViT-Huge (ViT-H, 636M parameters). We evaluated the two extreme variants---ViT-B representing a lightweight, computationally efficient option suitable for deployment on standard hardware, and ViT-H representing the most powerful but resource-intensive configuration---to establish performance bounds for CNT segmentation tasks. Both architectures share the same prompt encoder and mask decoder components but differ substantially in their image encoding capacity, with ViT-H providing approximately 7$\times$ more parameters for visual feature extraction.

We evaluated the segmentation performance of these two SAM architectures on carbon nanotube images acquired from both scanning electron microscopy and transmission electron microscopy platforms. The results are presented in Table \ref{tab:segmentation_results}. Both models demonstrated excellent segmentation accuracy across imaging modalities, with Dice coefficients exceeding 0.95 in all conditions. The ViT-H architecture consistently outperformed ViT-B across both imaging modalities, achieving Dice coefficients of 0.977 for SEM images and 0.982 for TEM images, compared to 0.954 and 0.970 for ViT-B, respectively. This superior performance of ViT-H was also reflected in IoU scores (0.957 and 0.965 for SEM and TEM) and precision metrics (0.983 and 0.985), demonstrating robust boundary delineation and reduced false positive segmentations. Notably, the ViT-H model exhibited lower variability in performance, particularly for SEM images, suggesting more consistent segmentation quality across diverse image characteristics.

Comparison across imaging modalities revealed that TEM images generally yielded higher segmentation accuracy than SEM images for both model architectures. Using the ViT-H model, TEM images achieved a marginally higher Dice coefficient (0.982) compared to SEM images (0.977), though this difference was modest. The performance gap between modalities was more pronounced with the ViT-B architecture, where TEM images (0.970) outperformed SEM images (0.954) by approximately 1.6 percentage points. This modality-dependent performance difference may be attributed to the higher contrast and resolution typically associated with TEM imaging, which provides clearer structural boundaries for segmentation. All models maintained accuracy values exceeding 99.8\%, indicating excellent pixel-level classification across the entire image field.

User effort analysis, quantified by the number of interactive clicks required to achieve satisfactory segmentation, revealed important practical considerations for model deployment. TEM images required substantially fewer user interventions than SEM images, with ViT-H requiring an average of 3.7 clicks for TEM compared to 6.4 clicks for SEM. Interestingly, while ViT-H demonstrated superior segmentation accuracy, it did not consistently reduce user effort compared to ViT-B, with the ViT-B model requiring slightly fewer clicks for SEM images (5.5 versus 6.4). The higher variability in click counts for SEM images suggests greater variability in image complexity or quality within this modality, necessitating more frequent manual refinement to achieve acceptable segmentation results. Overall, more than 90\% of images across all conditions achieved excellent segmentation performance (Dice $\geq$ 0.9), demonstrating the robust applicability of SAM models for automated carbon nanotube analysis in electron microscopy workflows.

Despite the theoretical advantages of the larger ViT-H architecture, performance gains over ViT-B were modest for CNT segmentation tasks, with Dice coefficient improvements of only 1.2--2.3 percentage points across imaging modalities. Given that ViT-B requires approximately 7$\times$ fewer parameters (91M versus 636M) while achieving near-equivalent segmentation quality, it represents the optimal choice for deployment in occupational health laboratories. The reduced memory footprint and faster inference times make ViT-B well-suited for routine use on standard desktop computers and laptops commonly found in field settings or resource-constrained environments. As a demonstration of accessibility, this entire workflow was developed on an M4 MacBook Pro with near-instantaneous inference performance, and should run efficiently on any modern laptop equipped with a GPU. For the classification experiments described in subsequent sections, we employed the ViT-B architecture as the feature encoder.

\begin{figure*}
    \centering
    \begin{subfigure}[b]{0.48\textwidth}
        \centering
        \includegraphics[width=\linewidth]{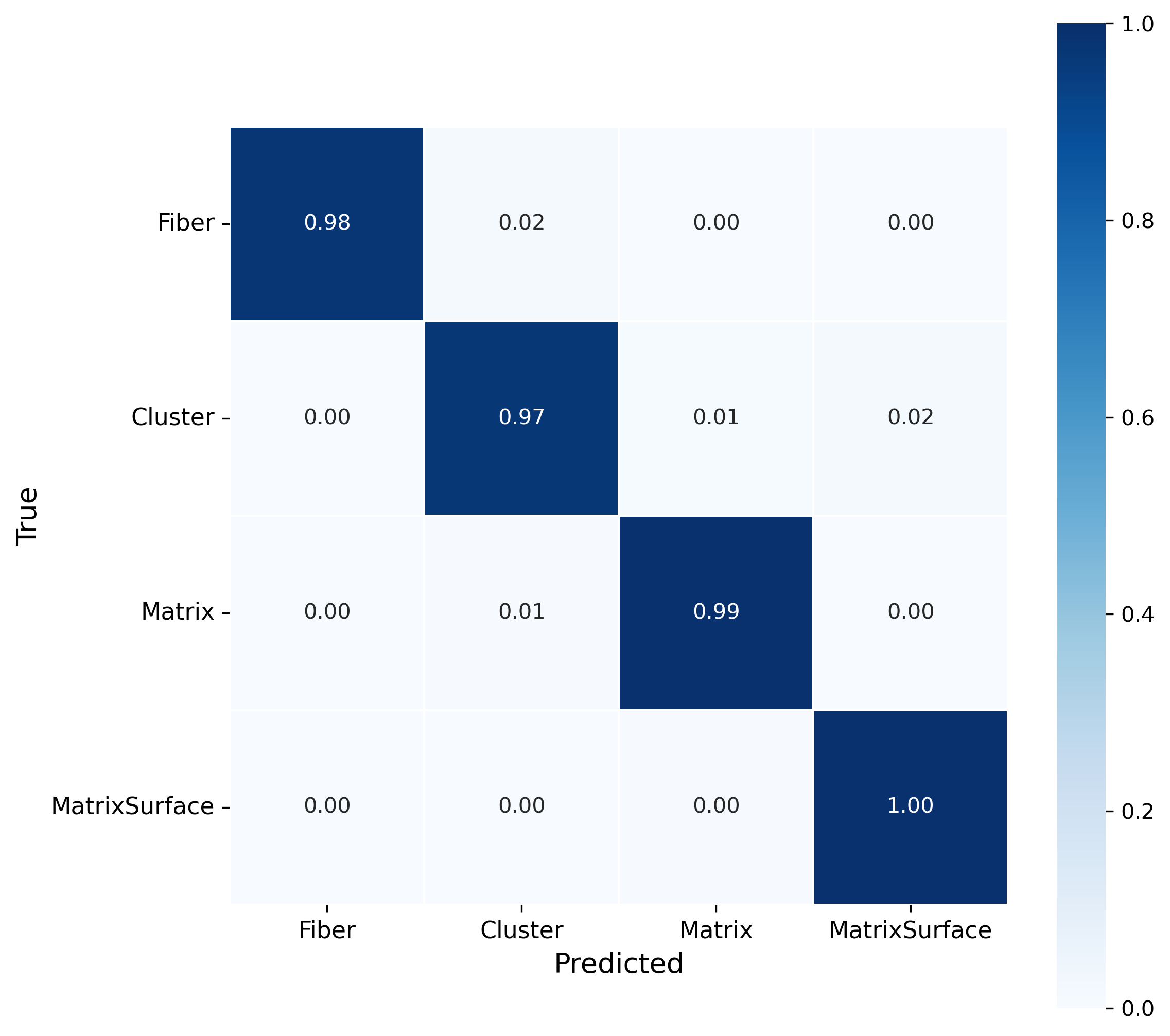}
        \vspace{0.2em}
        
        \centerline{\textbf{(a)} Entire Dataset}
        \label{fig:cm_all}
    \end{subfigure}
    \hfill 
    \begin{subfigure}[b]{0.48\textwidth}
        \centering
        \includegraphics[width=\linewidth]{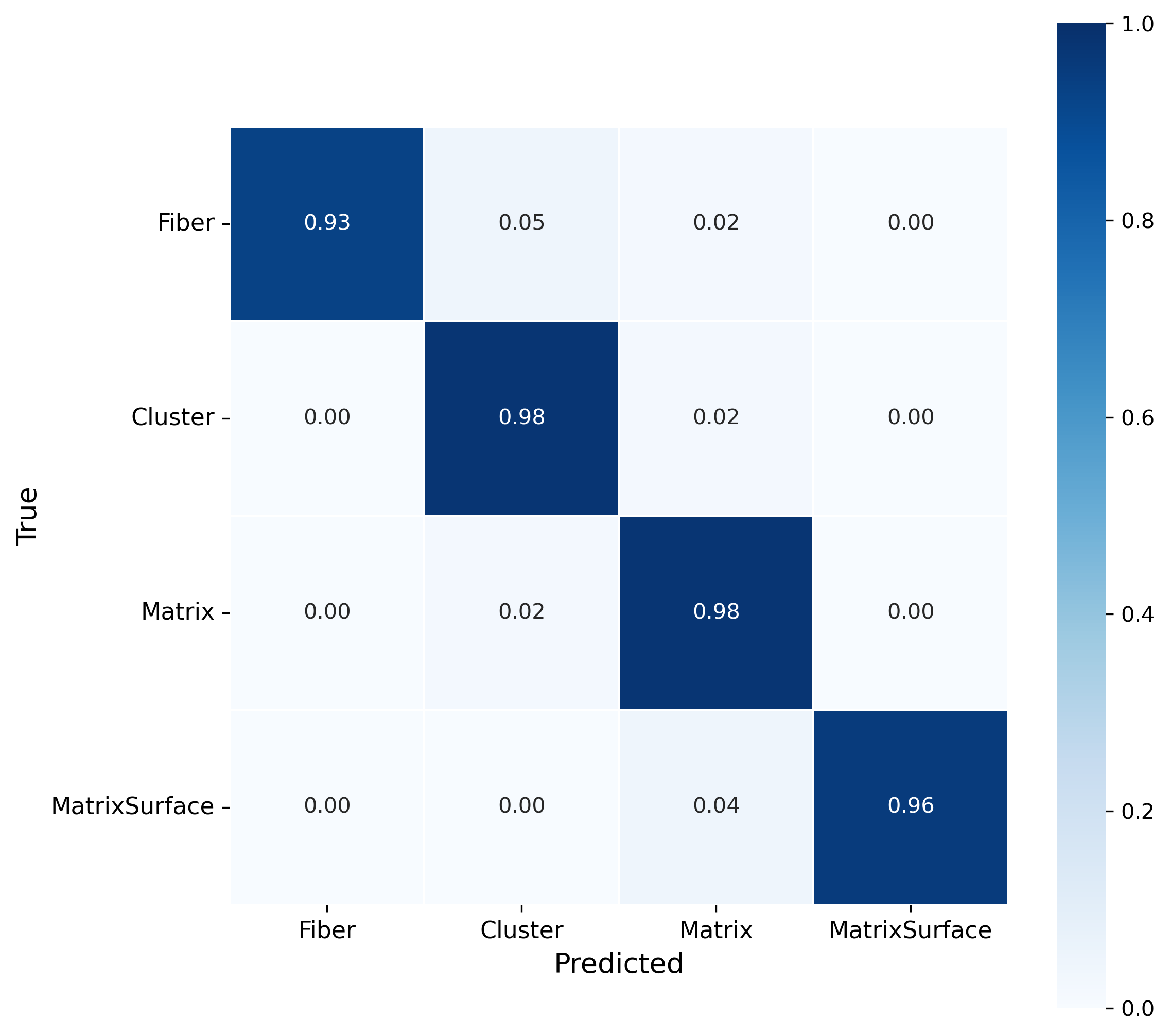}
        \vspace{0.2em}
        
        \centerline{\textbf{(b)} Held-out Test Set}
        \label{fig:cm_test}
    \end{subfigure}
    
    \caption{Comparative classification performance of the best-performing model. 
    \textbf{(a)} Normalized confusion matrix evaluated on the entire dataset, achieving 98.5\% overall accuracy. The strong diagonal demonstrates robust feature learning, with Matrix and MatrixSurface achieving near-perfect accuracy. Minor confusions occur primarily between morphologically adjacent classes: Fiber-to-Cluster (2\%) and Cluster-to-MatrixSurface (2\%). 
    \textbf{(b)} Normalized confusion matrix evaluated on the held-out test set, achieving 95.5\% overall accuracy. While Cluster (98\%) and Matrix (98\%) maintain strong performance, Fiber accuracy decreases to 93\% with 5\% misclassified as Clusters and 2\% as Matrix. MatrixSurface achieves 96\% accuracy with 4\% confused with Matrix. These test set errors reflect genuine morphological ambiguity in distinguishing overlapping fibers from loose bundles and in differentiating dense matrix regions from matrix-surface transitions.}
    \label{fig:confusion_matrices_combined}
\end{figure*}

\subsection{Classification Performance}

We systematically evaluated 24 model configurations combining two foundation models (SAM ViT-B/16 and DINOv2 ViT-B/14), two sampling strategies (mask-guided and uniform), three pooling methods (average, max, and avg+max concatenation), and two classifier types (linear and MLP). Test accuracy across all configurations ranged from 86.6\% to 95.5\%, with a mean of 91.4\% $\pm$ 2.4\% (Table~\ref{tab:model_comparison}; complete results in ESI Table 5). The best-performing architecture---DINOv2 with mask-guided sampling, avg+max pooling, and MLP classifier---achieved 95.5\% test accuracy, substantially outperforming the baseline method of Luo et al.\cite{Luo_Wang} (90.9\%) despite using only 34\% of their training data (1,620 vs 4,800 images). This improvement demonstrates that foundation models pre-trained on large-scale vision datasets can effectively transfer to specialized nanomaterial classification tasks with relatively small labeled datasets, and that segmentation-guided feature extraction provides benefits over uniform sampling approaches.

\begin{table}[b!]
\centering
\caption{Classification performance of selected model configurations. Models are grouped by encoder type and mask usage, with pooling strategy and classifier architecture as configuration parameters. All metrics computed on the held-out test set.}
\label{tab:model_comparison}
\small  
\setlength{\tabcolsep}{4pt}  
\begin{tabular}{@{}llllcc@{}}
\toprule
\textbf{Encoder} & \textbf{Mask} & \textbf{Pool} & \textbf{Classifier} & \textbf{Test Acc (\%)} & \textbf{Test F1} \\
\midrule
\multirow{3}{*}{DINOv2} & \multirow{3}{*}{Yes} 
    & A+M & MLP    & \textbf{95.53} & \textbf{0.956} \\
    &     & Avg & MLP    & 94.97 & 0.951 \\
    &     & Avg & Linear & 92.74 & 0.929 \\
\cmidrule(lr){2-6}
    & No  & Avg & MLP    & 93.85 & 0.939 \\
\midrule
\multirow{3}{*}{SAM} & \multirow{3}{*}{Yes} 
    & A+M & MLP    & 92.18 & 0.922 \\
    &     & Avg & MLP    & 91.62 & 0.918 \\
    &     & Avg & Linear & 91.62 & 0.916 \\
\cmidrule(lr){2-6}
    & No  & Avg & MLP    & 88.83 & 0.888 \\
\bottomrule
\multicolumn{6}{@{}l@{}}{\footnotesize A+M: Avg+Max pooling; Avg: Average pooling} \\
\end{tabular}
\end{table}

Analysis of the systematic experiments reveals several key insights. DINOv2 consistently outperformed SAM across all conditions (92.6\% vs 90.1\% mean accuracy) as expected, because self-supervised pre-training on diverse natural images yields more transferable representations than segmentation-specific pre-training for this classification task. Mask-guided sampling provided a consistent advantage over uniform sampling (92.7\% vs 90.1\%), with the largest improvements observed for DINOv2 (93.3\% masked vs 91.0\% uniform), indicating that focusing feature extraction on CNT regions effectively suppresses background noise and improves classification performance. Notably, however, maskless approaches still achieved strong performance, particularly with DINOv2, because the majority of images in this dataset contain a single dominant CNT morphology occupying most of the field of view. DINOv2's semantic understanding allows it to identify and focus on the relevant particle structures even when processing the entire image uniformly, though mask guidance still provides consistent improvement by eliminating ambiguity in multi-particle scenes or images with substantial background regions. Among pooling strategies, avg+max concatenation achieved the highest mean accuracy (91.8\%), slightly outperforming average pooling alone (91.6\%) and max pooling (90.6\%), suggesting that combining both typical and extreme feature activations captures complementary information about particle morphology. MLP classifiers consistently outperformed linear classifiers (92.0\% vs 90.8\%), demonstrating that shallow non-linear transformations can better exploit the rich feature representations produced by foundation models.

\begin{figure}[t!]
    \centering
    \includegraphics[width=\columnwidth]{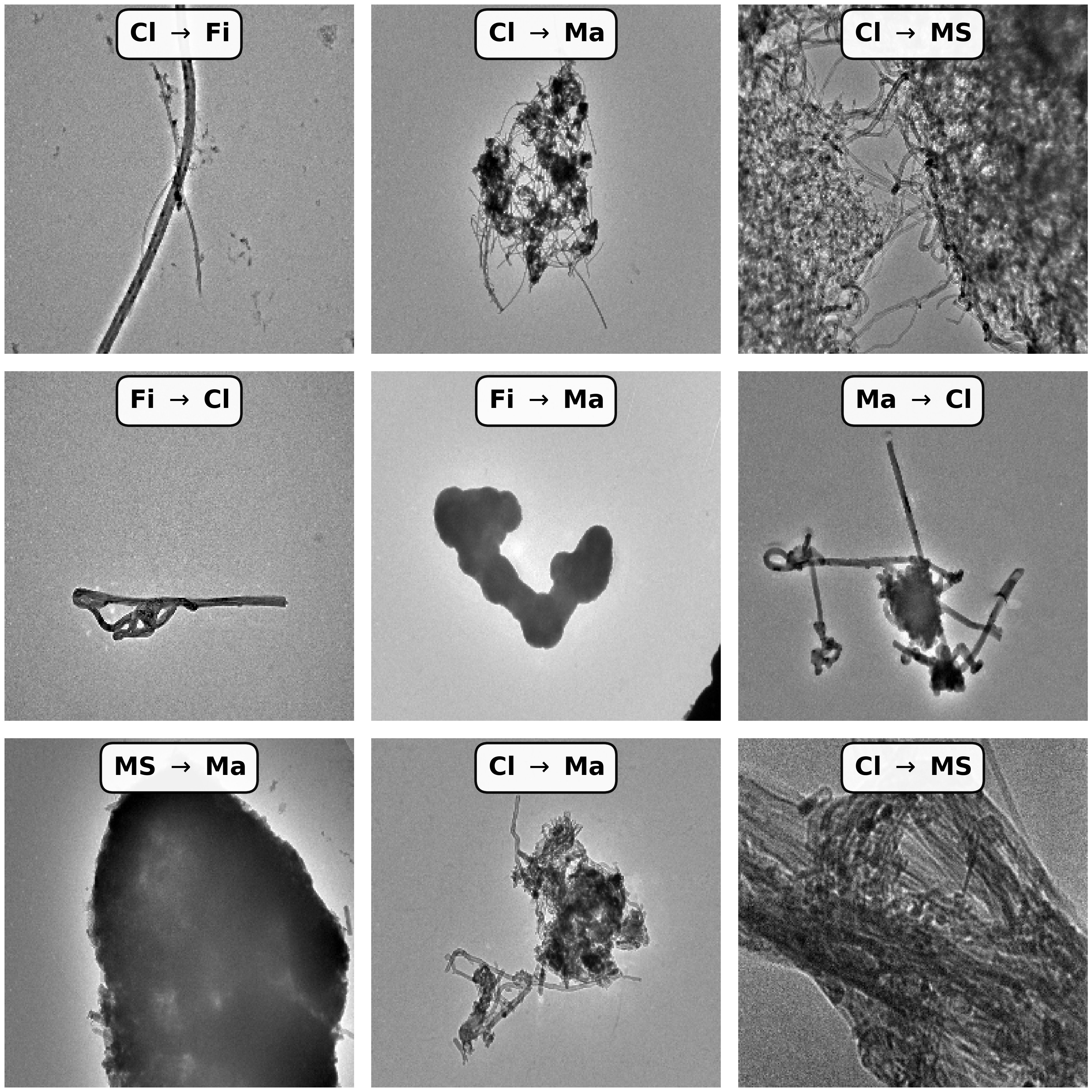}
    \caption{Representative examples of misclassified CNT images from the test set. Labels indicate true class $\rightarrow$ predicted class using abbreviated notation: Fi (Fiber), Cl (Cluster), Ma (Matrix), MS (MatrixSurface). These cases represent genuine morphological ambiguity at class boundaries where CNT structures exhibit transitional characteristics that challenge both automated and human expert classification.}
    \label{fig:misclassified_mosaic}
\end{figure}

\begin{figure}[b!]
    \centering
    \includegraphics[width=\linewidth]{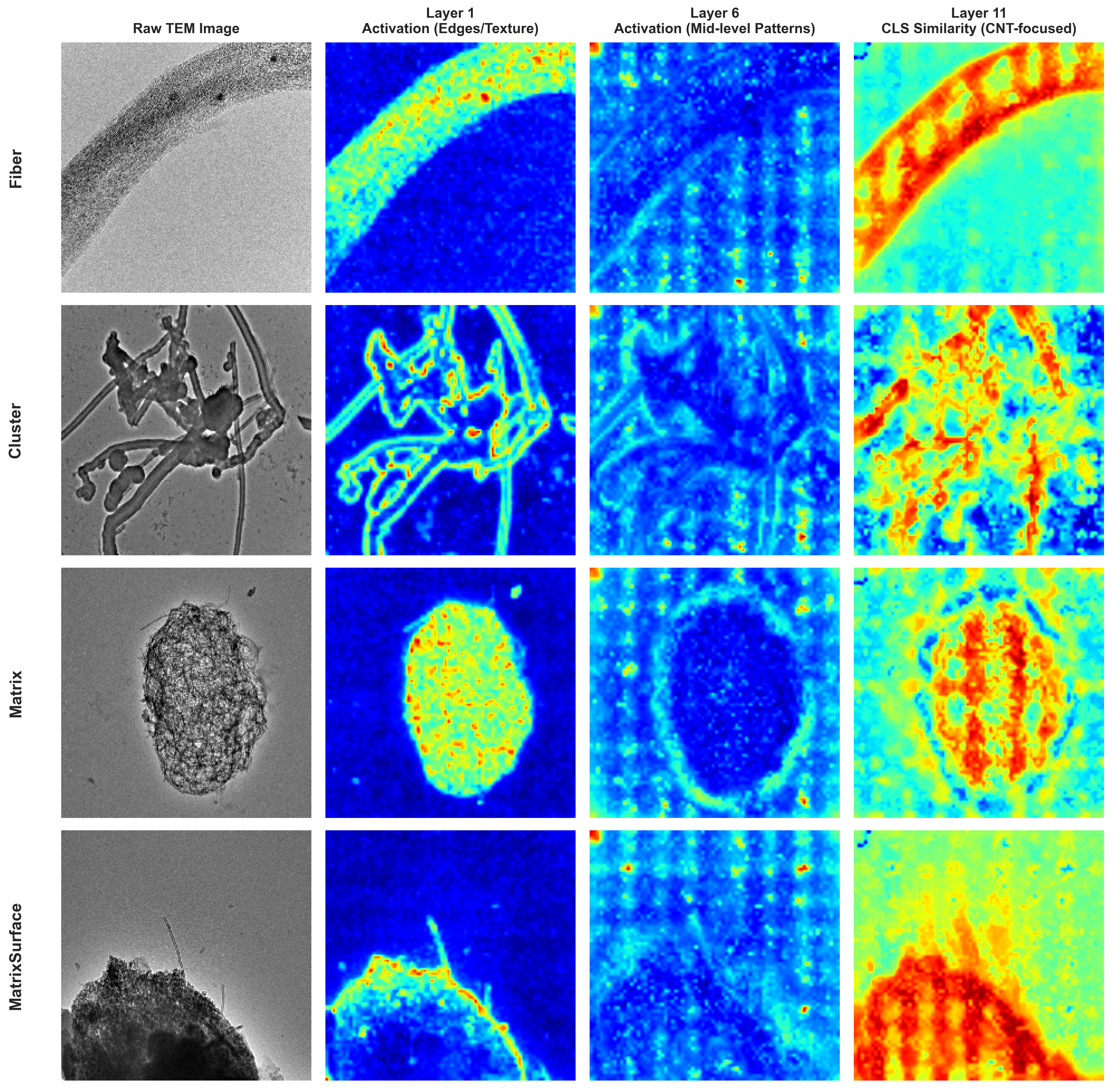}
    \caption{{How DINOv2 learns to recognize different CNT structures.} Each row shows one type of carbon nanotube morphology: individual fibers (top), loosely bundled clusters, densely packed matrices, and surface-embedded structures (bottom). The columns show how DINOv2's vision transformer processes these images at three different depths in the neural network. Early layers (Layer 1) detect simple features like edges and textures - notice the bright yellow-green activation along fiber boundaries and particle edges. Deep layers (Layer 11) develop a semantic understanding focused specifically on CNT regions, shown by the strong red activation on particles while ignoring the background. All heatmaps use blue for low activation and red for high activation.}
    \label{fig:activation_map}   
\end{figure}

\begin{figure*}
    \centering
    \includegraphics[width=\textwidth]{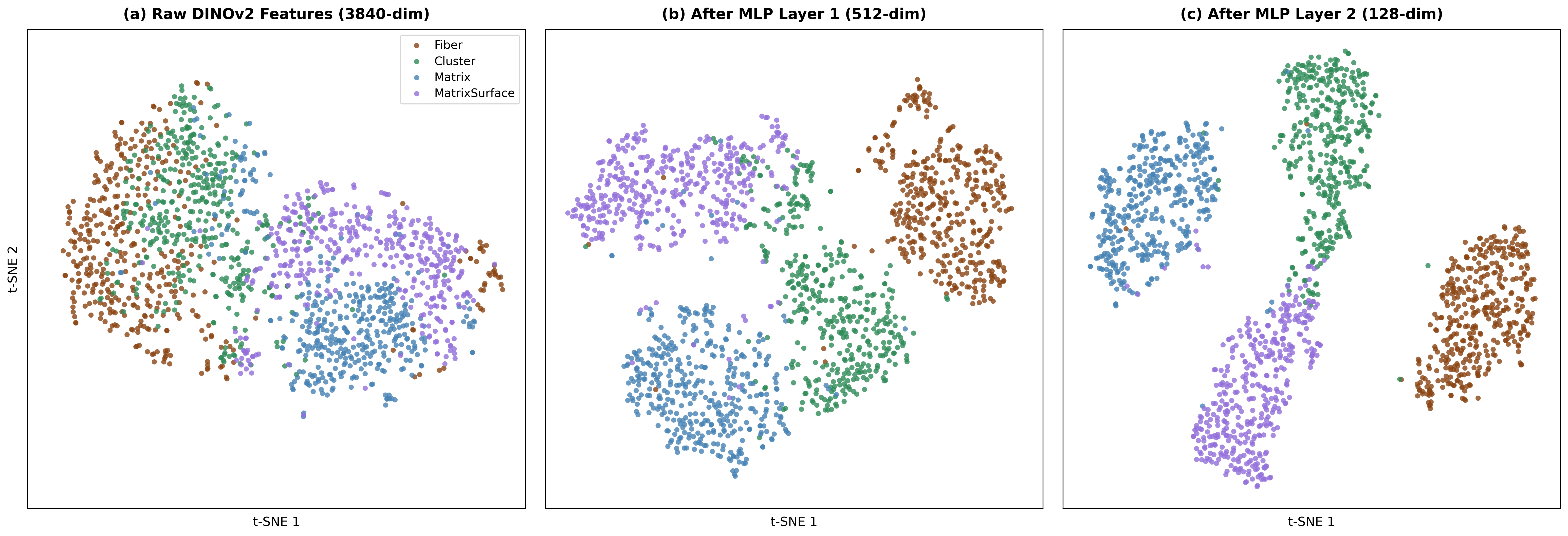}
    \caption{
        \textbf{t-SNE visualization of feature space evolution across the DINOv2 + MLP pipeline.}
        Each panel shows a two-dimensional t-SNE embedding of 400 randomly selected samples per class,
        colored by category (Fiber, Cluster, Matrix, MatrixSurface). Only the leftmost plot displays
        the $t$-SNE\,2 axis label for clarity.
        \textbf{(a)} Raw 3840-dimensional DINOv2 hypercolumn features before classification exhibit
        moderate separation by morphology.
        \textbf{(b)} After the first MLP layer (512-dimensional), class clusters become more compact and separable.
        \textbf{(c)} After the second MLP layer (128-dimensional), clusters form distinct, isolated manifolds
        corresponding to each CNT morphology, demonstrating that the MLP effectively learns
        class-specific embeddings from DINOv2 features.
        A small number of outlier points are visible in each class cluster, corresponding to images
        that were misclassified by the final model, highlighting morphological overlap
        across some difficult images.
    }
    \label{fig:tsne_dinov2_mlp_evolution}
\end{figure*}

The confusion matrices for our best-performing model (Figure~\ref{fig:confusion_matrices_combined}) reveal strong classification performance with interpretable error patterns reflecting genuine morphological ambiguity rather than systematic model failures. Evaluated across the entire dataset (Figure~\ref{fig:confusion_matrices_combined}a), the model achieves 98.5\% overall accuracy with MatrixSurface reaching near-perfect classification (99.6\%) and Matrix achieving 99\% accuracy, demonstrating that DINOv2's self-supervised features robustly capture the dense matrix structure and characteristic surface fiber extrusions that define these classes. Cluster and Fiber classes achieve 97\% and 98\% accuracy respectively, with structured misclassifications occurring primarily between morphologically adjacent categories: 2\% Fiber-to-Cluster confusion when multiple overlapping fibers obscure individual nanotube separation, and 2\% Cluster-to-MatrixSurface confusion when zoomed in images of tightly packed clusters resemble surface structures. The held-out test set (Figure~\ref{fig:confusion_matrices_combined}b) achieves 95.53\% overall accuracy, providing a more realistic assessment of generalization performance on unseen data. Fiber accuracy decreases to 93\%, with 5\% misclassified as Cluster when fiber entanglement or close spatial proximity creates ambiguous bundle-like appearances, and 2\% as Matrix when densely packed orientations obscure individual fiber separation. Cluster (98\%) and Matrix (98\%) maintain robust performance, while MatrixSurface accuracy decreases to 96\% with 4\% confusion with Matrix when surface extrusions are absent. Visual examination of representative misclassified examples (Figure~\ref{fig:misclassified_mosaic}) confirms that Cluster errors dominate the error distribution, reflecting this transitional morphology's inherent difficulty: loosely bundled clusters (Cl→Fi) resemble isolated fibers, densely packed clusters (Cl→Ma) approach matrix-like compaction, and clusters with surface structures (Cl→MS) exhibit MatrixSurface characteristics. Matrix-to-Cluster confusion arises when individual fiber structures remain distinguishable within the matrix, while Fiber errors occur when multiple overlapping nanotubes create bundle-like appearances. Critically, there is very little confusion between structurally dissimilar classes (e.g., Fiber-to-Matrix or Matrix-to-Fiber), validating that the model learns meaningful structural representations rather than superficial image artifacts.

To understand why DINOv2 substantially outperforms both traditional CNNs and SAM for CNT classification, we visualized the hierarchical feature representations learned by the model across different transformer layers (Figure \ref{fig:activation_map}). These activation maps reveal how DINOv2 progressively builds increasingly abstract and task-specific representations of CNT morphologies \cite{raghu2021vision,hariharan2015hypercolumns}. Layer 1 activations respond primarily to low-level visual features---edges, texture gradients, and local contrast variations---producing bright activation along fiber boundaries, particle edges, and regions of high structural detail. These early-layer features are largely task-agnostic, capturing primitive visual elements present in any natural or microscopy image. By Layer 6, these distinct spatial boundaries begin to diffuse as the transformer integrates information across image patches; the resulting grid-like patterns indicate that the model is mixing local textures into distributed mid-level representations rather than encoding explicit object boundaries at this stage. The most striking transformation occurs in Layer 11, where we shift from visualizing raw activations to visualizing CLS token similarity - a proxy for the model's "visual attention". These maps show that the model recovers spatial specificity but with semantic meaning: it activates strongly and exclusively on CNT particles while actively suppressing background regions, and imaging artifacts like TEM grid bars. This selective focus emerges naturally from self-supervised pre-training, confirming the emergent segmentation properties previously observed in DINO-based architectures\cite{caron2021emerging,dosovitskiy2020image}, which effectively learn to "look" only at the relevant nanomaterial even without explicit segmentation training.

These visualization results directly explain our quantitative findings that mask-guided sampling provides consistent but modest improvements over uniform sampling. Because DINOv2's deep layers already learn to focus on task-relevant particle regions as evidenced by the selective Layer 11 activation maps - the model partially performs its own implicit "segmentation" during feature learning. Mask-guided sampling still improves performance by eliminating background tokens entirely during feature aggregation, reducing noise in the pooled representations and allowing the model to dedicate more representational capacity to within-particle variations that distinguish the four morphology classes. However, the benefit is less dramatic than it would be for architectures without this learned selectivity. In contrast, SAM shows a larger performance gap between masked and uniform sampling conditions, consistent with SAM's segmentation-focused pre-training that may not inherently prioritize semantic discriminability between particle types. The hierarchical feature progression visualized here---from generic edge detection to task-specific semantic understanding---also explains why concatenating features from multiple layers outperforms single-scale approaches: early layers provide fine-grained textural detail necessary for distinguishing subtle morphological differences, while late layers contribute discriminative, class-specific information that suppresses confusing background variations.

To visualize how the MLP classifier transforms DINOv2's foundation model features into highly discriminative representations, we applied t-SNE \cite{VanDerMaaten_Hinton_TSNE_2008,VanDerMaaten_BarnesHutSNE_2013,VanDerMaaten_AcceleratingTSNE_2014} dimensionality reduction to embeddings at three stages of the classification pipeline (Figure \ref{fig:tsne_dinov2_mlp_evolution}). The raw 3840-dimensional DINOv2 hypercolumn features (Figure \ref{fig:tsne_dinov2_mlp_evolution}a) already exhibit meaningful structure: the four CNT morphology classes form loosely separated clusters, with Fiber and Matrix samples occupying relatively distinct regions while Fiber and Cluster samples show greater overlap, consistent with their morphological similarity in a few images. This initial separation validates that DINOv2's self-supervised pre-training on natural images transfers effectively to nanomaterial microscopy, capturing morphology-relevant features without task-specific fine-tuning. However, substantial inter-class overlap remains, particularly between Fiber and Cluster samples, and between Matrix and MatrixSurface samples explaining the 2\% classification error in the confusion matrix. The first MLP layer (512 dimensions, Figure \ref{fig:tsne_dinov2_mlp_evolution}b) significantly refines this structure: class clusters become more compact and better separated, with decision boundaries beginning to form between previously overlapping regions. The most dramatic transformation occurs after the second MLP layer (128 dimensions, \ref{fig:tsne_dinov2_mlp_evolution}c), where all four classes form tight, well-separated manifolds with minimal overlap. This progression demonstrates that the MLP learns to amplify task-relevant dimensions of the DINOv2 feature space while suppressing irrelevant variations, effectively discovering a low-dimensional embedding optimized for CNT morphology discrimination. The small number of outlier points visible in each cluster correspond precisely to the misclassified samples in the confusion matrix, representing genuinely ambiguous borderline cases rather than random errors. Notably, even in the final 128-dimensional space, MatrixSurface and Cluster outliers tend to drift toward each other's clusters, while Fiber and Matrix remain distinctly separated, confirming that the 2\% Cluster-to-MatrixSurface confusion reflects true morphological ambiguity along a continuum rather than model failure. This visualization provides interpretable evidence that our model's 95.5\% test accuracy stems from learning a geometrically meaningful feature space where morphological similarity corresponds directly to proximity in the embedding, rather than memorizing superficial image patterns.

\begin{figure}[t]
    \centering
    \includegraphics[width=\columnwidth]{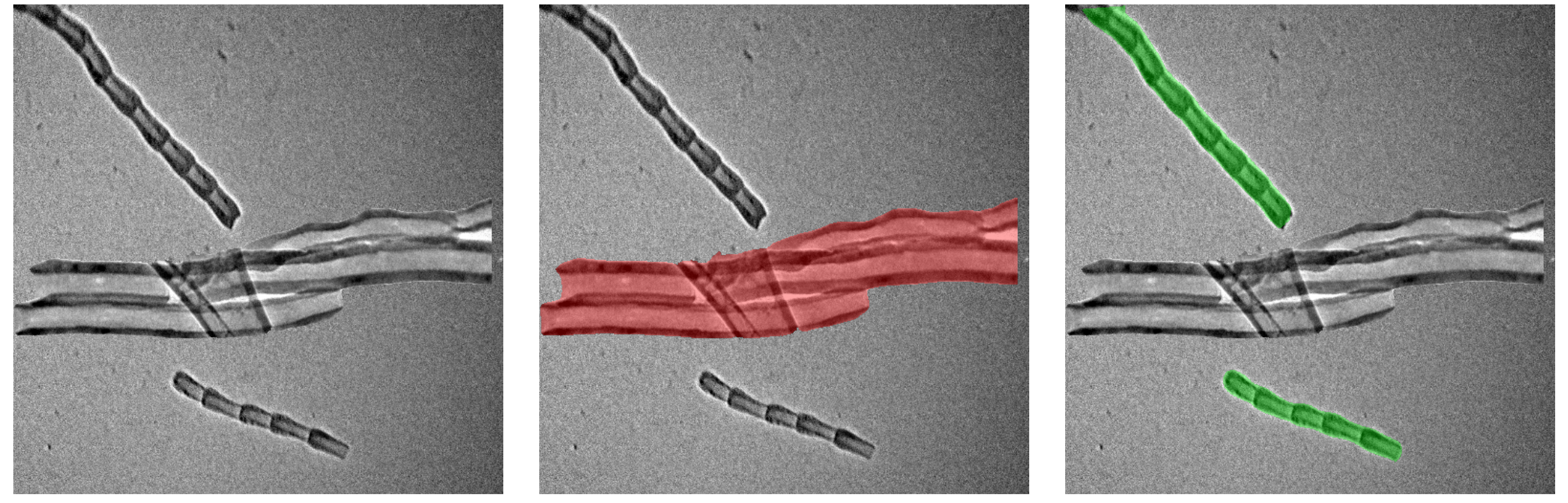}\\[0.1em]
    \includegraphics[width=\columnwidth]{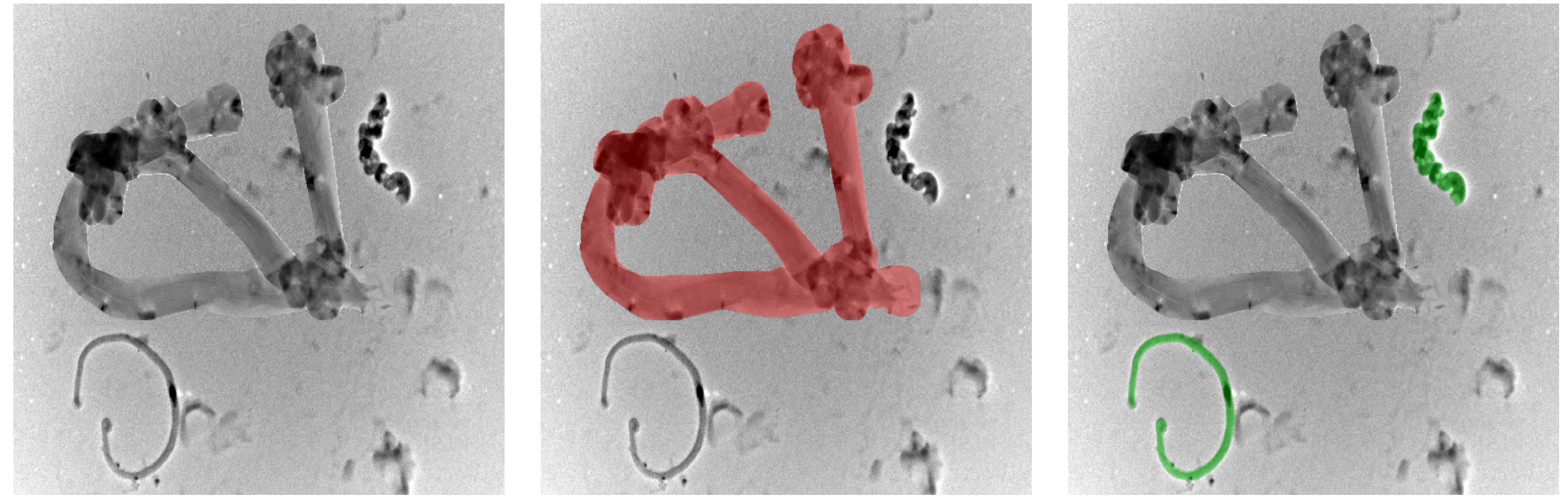}\\[0.1em]
    \includegraphics[width=\columnwidth]{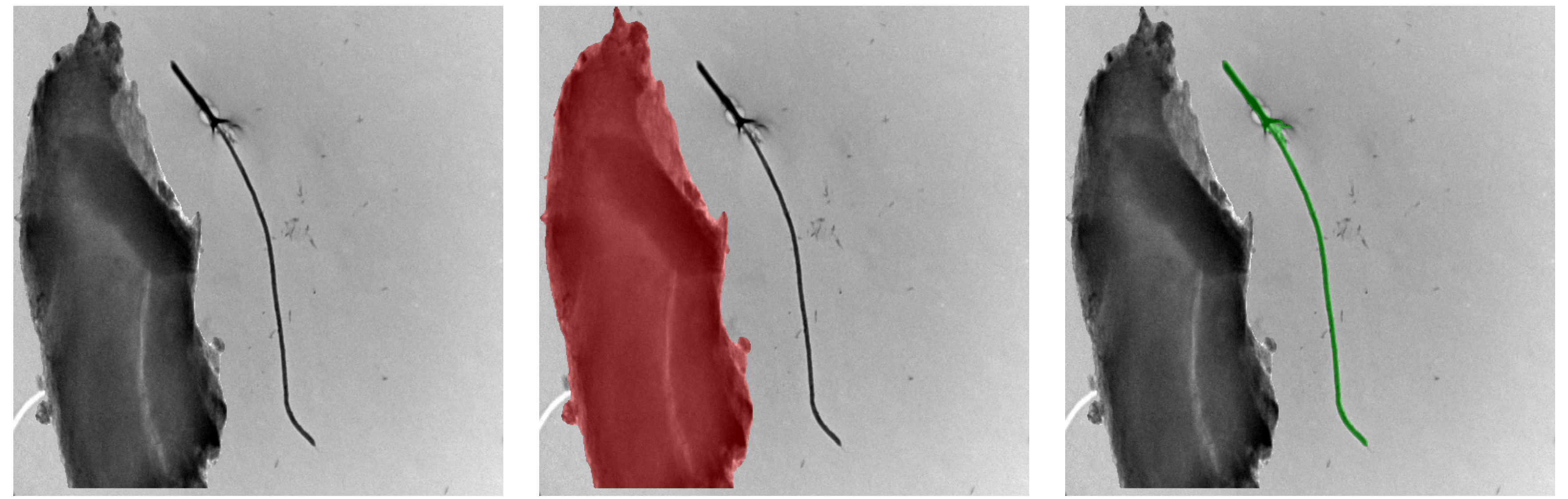}\\[0.1em]
    \includegraphics[width=\columnwidth]{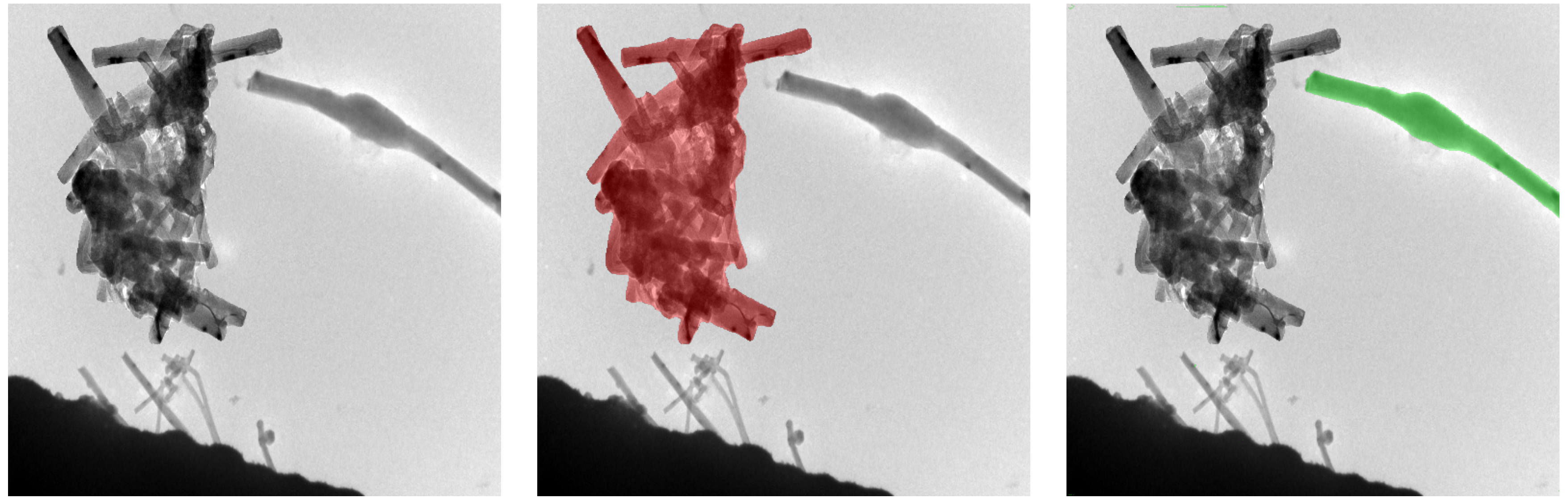}\\[0.3em]
    \makebox[\columnwidth]{%
        \makebox[0.33\columnwidth]{\textbf{(a)}}\hfill
        \makebox[0.33\columnwidth]{\textbf{(b)}}\hfill
        \makebox[0.33\columnwidth]{\textbf{(c)}}%
    }
    \caption{\small\sffamily{Mask-guided multi-particle classification on synthetic composite images.} 
    (a) Composite images created by digitally inserting CNT particles from different morphological classes into the same field of view. 
    (b) Target particles (Cluster in rows 1--2, Matrix in rows 3--4) correctly identified using red mask overlays. 
    (c) Co-existing Fiber particles correctly identified using green mask overlays. 
    The mask guidance enables selective classification of individual particles within multi-particle scenes, demonstrating the model's ability to analyze spatially separated structures.}
    \label{fig:mask_guided_demo}
\end{figure}

To validate the robustness of our mask-guided classification approach for multi-particle images, we created synthetic composite images by digitally copying particles from one morphological class and placing them alongside particles from another class within the same field of view. This controlled experimental design allows us to assess whether the classifier can accurately identify each particle independently when multiple CNT structures with different morphologies are present in a single image. Figure~\ref{fig:mask_guided_demo} demonstrates representative examples where the model correctly classifies Cluster, Matrix, and Fiber particles within these composite scenes using appropriate mask guidance for each structure. Preliminary testing on these synthetic composites indicates classification accuracy exceeding 85\%, confirming that mask-guided feature sampling effectively isolates morphological information from individual particles. However, we note that this approach is sensitive to DINOv2's feature map resolution of $37 \times 37$ pixels; particles in close spatial proximity may exhibit some degree of feature overlap or mutual influence at this scale, potentially affecting classification boundaries. For images containing densely packed particle fields, a crop-based workflow---where individual segmented particles are extracted and processed independently at higher effective resolution---may provide more robust classification by eliminating inter-particle interference and ensuring that feature extraction focuses exclusively on a single structure at a time.

\section{Conclusions}
This work demonstrates that foundation models pre-trained on natural images can be effectively adapted for automated carbon nanotube characterization in electron microscopy, achieving 98.5\% classification accuracy across the entire dataset and 95.5\% on held-out test data. Our two-stage pipeline combines the Segment Anything Model (SAM) for particle segmentation with DINOv2 vision transformers for morphology-based classification, substantially outperforming previous CNN-based approaches despite using only one-third of the training data. The key insight enabling this performance is mask-guided feature extraction: by using SAM-generated segmentation masks to spatially constrain DINOv2's attention to particle regions, we effectively suppress background noise and focus the model's representational capacity on morphologically relevant structural variations. Systematic ablation across 24 model configurations revealed that DINOv2 consistently outperforms SAM for classification (92.6\% vs 90.1\% mean accuracy), that mask-guided sampling provides consistent advantages over uniform sampling (92.7\% vs 90.1\%), and that shallow MLP classifiers better exploit foundation model features than linear heads (92.0\% vs 90.8\%). Visualization of learned representations through activation maps and t-SNE embeddings confirms that the model captures semantically meaningful structural hierarchies, from low-level edge detection to high-level morphological concepts, explaining both its strong performance and its interpretable error patterns concentrated at genuine morphological boundaries between particle classes.

While our framework targets CNTs, the integration of foundation models for automated segmentation and morphological classification addresses a broader challenge in nanomaterial characterization: manual electron microscopy analysis is inherently slow, subjective, and incompatible with modern high-throughput materials synthesis that requires rapid quality assessment. Traditional workflows demand hours of expert time to analyze even modest sample sizes, preventing systematic quantification of morphological distributions and making it impossible to provide real-time feedback during production. Our SAM-guided DINOv2 approach demonstrates that large-scale pre-trained vision models can be efficiently adapted to diverse nanoparticle systems, enabling rapid extraction of morphological parameters such as size, shape, and aggregation state at scales previously requiring extensive manual effort. This automation makes it feasible to integrate electron microscopy into routine manufacturing workflows, where immediate morphological feedback can guide process adjustments to achieve desired particle characteristics, transforming EM from a post-synthesis analytical tool into an active component of materials optimization.

Several promising extensions could further enhance the framework's capabilities and applicability. First, incorporating classical morphological features such as aspect ratio, circularity, solidity, texture based features and Hu moments, alongside DINOv2 embeddings could improve discrimination of particles with subtle shape differences\cite{mewada2025hybrid}, particularly for distinguishing loosely bundled clusters from compact matrix structures. Such hybrid feature representations would combine the semantic understanding of foundation models with explicit geometric constraints validated across decades of materials characterization. Second, the current mask-guided approach assumes that features within a masked region belong to a single morphological class, which becomes problematic when different particle types overlap. In such cases, the extracted features contain mixed signals from multiple morphologies, potentially confusing the classifier. To resolve this, future work could incorporate instance segmentation to detect and individually crop overlapping particles \cite{frei2023deep,kraus2016classifying}. By upscaling and treating each crop as a standalone input, DINOv2 could classify the distinct morphology of each component independently, enabling robust analysis even in crowded fields. Third, integrating multi-modal data such as combining morphological classification from TEM with elemental composition from energy-dispersive X-ray spectroscopy (EDS) or structural information from electron diffraction, could provide more comprehensive particle characterization for complex engineered nanomaterials\cite{adak2022imaging}.

The most impactful future direction involves retraining foundation models on large-scale, domain-specific electron microscopy datasets spanning diverse nanomaterial types, imaging conditions, and morphological variations. While transfer learning from natural image pre-training proves surprisingly effective, models trained explicitly on a large number of annotated EM images would likely capture nanoscale structural motifs, microscopy-specific artifacts, and fine morphological distinctions that current general-purpose vision models cannot reliably encode. Such domain-specialized foundation models---analogous to how medical imaging has developed radiology-specific vision transformers\cite{shamshad2022transformersmedicalimagingsurvey}---would represent a critical step toward fully automated quantification and classification of complex-shaped nanomaterials including CNTs, CNFs, metal oxide nanoparticles, graphene, and hybrid nanocomposites. Combined with advances in automated microscopy hardware and real-time image acquisition, these AI-enabled characterization pipelines could fundamentally transform nanomaterial quantification and classification from an art requiring expert manual oversight into a data-driven science.

\section*{Author contributions}
S.P. developed the methodology, performed all computational experiments, and wrote the manuscript. C.W. performed TEM analysis and image annotation. M.M.D. collected field samples. J.D.E. provided supervision and manuscript review. C.S.J.T. provided domain expertise and manuscript review.

\section*{Conflicts of interest}
The findings and conclusions in this report are those of the  authors and do not necessarily represent the views of the National Institute for Occupational Safety and Health. Mention of product or company name does not constitute endorsement by the Centers for Disease Control and Prevention.

\section*{Data availability}
The code for the interactive segmentation tool and classification pipeline is available at \url{https://github.com/SanjayPradeep97/SAM-SEM-Segmentation}. The datasets used in this study are publicly available at the Harvard Dataverse repository: \url{https://doi.org/10.7910/DVN/5O0SF7}.

\section*{Acknowledgements}
This research was supported by the National Institute for Occupational Safety and Health (NIOSH) of the Centers for Disease Control and Prevention (CDC) under award number R21OH012397. We gratefully acknowledge Joe Fernback at NIOSH for pioneering microscopy work that initiated the dataset, and Hope Davey, Amelia Chen, and Judy Su at UCLA for sample collection and analysis.


\balance

\bibliography{references}

@article{Luo_Wang,
  author  = {Luo, Qixiang and Holm, Elizabeth A. and Wang, Chen},
  title   = {A transfer learning approach for improved classification of carbon nanomaterials from TEM images},
  journal = {Nanoscale Adv.},
  year    = {2021},
  volume  = {3},
  number  = {1},
  pages   = {206--213},
  doi     = {10.1039/D0NA00634C},
  url     = {http://dx.doi.org/10.1039/D0NA00634C},
}

@data{HarvardDataverse,
author = {Wang, Chen},
publisher = {Harvard Dataverse},
title = {{Dataset of TEM Images for Carbon Nanomaterial Classification}},
year = {2024},
version = {V1},
doi = {10.7910/DVN/5O0SF7},
url = {https://doi.org/10.7910/DVN/5O0SF7}
}

@article{KatherineDuke,
author = {Duke, Katherine S. and Bonner, James C.},
title = {Mechanisms of carbon nanotube-induced pulmonary fibrosis: a physicochemical characteristic perspective},
journal = {WIREs Nanomedicine and Nanobiotechnology},
volume = {10},
number = {3},
pages = {e1498},
doi = {https://doi.org/10.1002/wnan.1498},
url = {https://wires.onlinelibrary.wiley.com/doi/abs/10.1002/wnan.1498},
eprint = {https://wires.onlinelibrary.wiley.com/doi/pdf/10.1002/wnan.1498},
year = {2018}
}

@article{Schulte2019-jv,
  title     = "Current state of knowledge on the health effects of engineered
               nanomaterials in workers: a systematic review of human studies
               and epidemiological investigations",
  author    = "Schulte, Paul A and Leso, Veruscka and Niang, Mamadou and
               Iavicoli, Ivo",
  journal   = "Scand. J. Work Environ. Health",
  publisher = "Scandinavian Journal of Work, Environment and Health",
  volume    =  45,
  number    =  3,
  pages     = "217--238",
  month     =  may,
  year      =  2019,
  language  = "en"
}

@article{SHVEDOVA2009192,
author = {A.A. Shvedova and E.R. Kisin and D. Porter and P. Schulte and V.E. Kagan and B. Fadeel and V. Castranova},
title = {Mechanisms of pulmonary toxicity and medical applications of carbon nanotubes: Two faces of Janus?},
journal = {Pharmacology \& Therapeutics},
volume = {121},
number = {2},
pages = {192-204},
year = {2009},
issn = {0163-7258},
doi = {https://doi.org/10.1016/j.pharmthera.2008.10.009},
url = {https://www.sciencedirect.com/science/article/pii/S0163725808002179}
}

@article{DONALDSON20132078,
author = {Ken Donaldson and Craig A. Poland and Fiona A. Murphy and Marion MacFarlane and Tatyana Chernova and Anja Schinwald},
title = {Pulmonary toxicity of carbon nanotubes and asbestos — Similarities and differences},
journal = {Advanced Drug Delivery Reviews},
volume = {65},
number = {15},
pages = {2078-2086},
year = {2013},
note = {Carbon Nanotubes in Medicine and Biology: Therapy and Diagnostics \& Safety and Toxicology},
issn = {0169-409X},
doi = {https://doi.org/10.1016/j.addr.2013.07.014},
url = {https://www.sciencedirect.com/science/article/pii/S0169409X13001683}
}

@article{Fiji,
  author    = {Johannes Schindelin and Ignacio Arganda-Carreras and Erwin Frise and Verena Kaynig and Mark Longair and Tobias Pietzsch and Stephan Preibisch and Curtis Rueden and Stephan Saalfeld and Benjamin Schmid and Jean-Yves Tinevez and Daniel J. White and Volker Hartenstein and Kevin Eliceiri and Pavel Tomancak and Albert Cardona},
  title     = {Fiji: an open-source platform for biological-image analysis},
  journal   = {Nature Methods},
  year      = {2012},
  volume    = {9},
  number    = {7},
  pages     = {676--682},
  doi       = {10.1038/nmeth.2019},
  url       = {https://doi.org/10.1038/nmeth.2019}
}

@article{ImageJ,
author = {Tony J. Collins},
title = {ImageJ for Microscopy},
journal = {BioTechniques},
volume = {43},
number = {sup1},
pages = {S25--S30},
year = {2007},
publisher = {Taylor \& Francis},
doi = {10.2144/000112517},

    note ={PMID: 17936939},


URL = { 
    
        https://doi.org/10.2144/000112517
    
    

},
eprint = { 
    
        https://doi.org/10.2144/000112517
    
    

}

}

@article{deep_microscopy4,
  title={Deep learning in electron microscopy},
  author={Ede, Jeffrey M},
  journal={Machine Learning: Science and Technology},
  volume={2},
  number={1},
  pages={011004},
  year={2021},
  publisher={IOP Publishing}
}

@article{deep_microscopy,
    author = {Treder, Kevin P and Huang, Chen and Kim, Judy S and Kirkland, Angus I},
    title = {Applications of deep learning in electron microscopy},
    journal = {Microscopy},
    volume = {71},
    number = {Supplement_1},
    pages = {i100-i115},
    year = {2022},
    month = {02},
    issn = {2050-5701},
    doi = {10.1093/jmicro/dfab043},
    url = {https://doi.org/10.1093/jmicro/dfab043},
    eprint = {https://academic.oup.com/jmicro/article-pdf/71/Supplement_1/i100/42548156/dfab043.pdf},
}

@Article{deep_microscopy2,
AUTHOR = {López Gutiérrez, Jorge David and Abundez Barrera, Itzel Maria and Torres Gómez, Nayely},
TITLE = {Nanoparticle Detection on SEM Images Using a Neural Network and Semi-Synthetic Training Data},
JOURNAL = {Nanomaterials},
VOLUME = {12},
YEAR = {2022},
NUMBER = {11},
ARTICLE-NUMBER = {1818},
ARTICLE-NUMBER = {1818},
pages = {n/a},
URL = {https://www.mdpi.com/2079-4991/12/11/1818},
PubMedID = {35683674},
ISSN = {2079-4991},
DOI = {10.3390/nano12111818}
}

@article{deep_microscopy3,
doi = {10.1088/2515-7639/ad229b},
url = {https://doi.org/10.1088/2515-7639/ad229b},
year = {2024},
month = {feb},
publisher = {IOP Publishing},
volume = {7},
number = {2},
pages = {022001},
author = {Chen, K and Barnard, A S},
title = {Advancing electron microscopy using deep learning},
journal = {Journal of Physics: Materials}
}

@Article{Trujillo2017,
author={Trujillo, Mar{\'i}a Celeste Ram{\'i}rez
and Alarc{\'o}n, Teresa E.
and Dalmau, Oscar S.
and Zamudio Ojeda, Adalberto},
title={Segmentation of carbon nanotube images through an artificial neural network},
journal={Soft Computing},
year={2017},
month={Feb},
day={01},
volume={21},
number={3},
pages={611-625},
issn={1433-7479},
doi={10.1007/s00500-016-2426-1},
url={https://doi.org/10.1007/s00500-016-2426-1}
}

@inproceedings{yolo,
  title={You only look once: Unified, real-time object detection},
  author={Redmon, Joseph and Divvala, Santosh and Girshick, Ross and Farhadi, Ali},
  booktitle={Proceedings of the IEEE conference on computer vision and pattern recognition},
  pages={779--788},
  year={2016}
}

@inproceedings{unet,
  title={U-net: Convolutional networks for biomedical image segmentation},
  author={Ronneberger, Olaf and Fischer, Philipp and Brox, Thomas},
  booktitle={International Conference on Medical image computing and computer-assisted intervention},
  pages={234--241},
  year={2015},
  organization={Springer}
}

@article{unet_EM,
title = {Automated image segmentation of scanning electron microscopy images of graphene using U-Net Neural Network},
journal = {Materials Today Communications},
volume = {35},
pages = {106127},
year = {2023},
issn = {2352-4928},
doi = {https://doi.org/10.1016/j.mtcomm.2023.106127},
url = {https://www.sciencedirect.com/science/article/pii/S2352492823008188},
author = {Aagam Shah and Joshua A. Schiller and Isiah Ramos and James Serrano and Darren K. Adams and Sameh Tawfick and Elif Ertekin}
}

@article{yolo_TEM,
title = {A deep learning based automatic defect analysis framework for In-situ TEM ion irradiations},
journal = {Computational Materials Science},
volume = {197},
pages = {110560},
year = {2021},
issn = {0927-0256},
doi = {https://doi.org/10.1016/j.commatsci.2021.110560},
url = {https://www.sciencedirect.com/science/article/pii/S0927025621002871},
author = {Mingren Shen and Guanzhao Li and Dongxia Wu and Yudai Yaguchi and Jack C. Haley and Kevin G. Field and Dane Morgan}
}

@article{yolo_drawback1,
author = {Rettenberger, Luca and Szymanski, Nathan and Zeng, Yan and Schuetzke, Jan and Wang, Shilong and Ceder, Gerbrand and Reischl, Markus},
year = {2024},
month = {06},
pages = {},
title = {Uncertainty-aware particle segmentation for electron microscopy at varied length scales},
volume = {10},
journal = {npj Computational Materials},
doi = {10.1038/s41524-024-01302-w}
}

@inproceedings{yolo_drawback2,
  title={Semi-supervised segmentation of mitochondria from electron microscopy images using spatial continuity},
  author={Xiao, Yunpeng and Zhao, Youpeng and Yang, Ge},
  booktitle={2022 IEEE 19th International Symposium on Biomedical Imaging (ISBI)},
  pages={1--5},
  year={2022},
  organization={IEEE}
}

@article{vgg,
  title={Very deep convolutional networks for large-scale image recognition},
  author={Simonyan, Karen and Zisserman, Andrew},
  journal={arXiv preprint arXiv:1409.1556},
  year={2014}
}

@inproceedings{arandjelovic2016netvlad,
  title={NetVLAD: CNN architecture for weakly supervised place recognition},
  author={Arandjelovic, Relja and Gronat, Petr and Torii, Akihiko and Pajdla, Tomas and Sivic, Josef},
  booktitle={Proceedings of the IEEE conference on computer vision and pattern recognition},
  pages={5297--5307},
  year={2016}
}

@inproceedings{kirillov2023segment,
  title={Segment anything},
  author={Kirillov, Alexander and Mintun, Eric and Ravi, Nikhila and Mao, Hanzi and Rolland, Chloe and Gustafson, Laura and Xiao, Tete and Whitehead, Spencer and Berg, Alexander C and Lo, Wan-Yen and others},
  booktitle={Proceedings of the IEEE/CVF international conference on computer vision},
  pages={4015--4026},
  year={2023}
}

@Article{Archit2025,
author={Archit, Anwai
and Freckmann, Luca
and Nair, Sushmita
and Khalid, Nabeel
and Hilt, Paul
and Rajashekar, Vikas
and Freitag, Marei
and Teuber, Carolin
and Spitzner, Melanie
and Tapia Contreras, Constanza
and Buckley, Genevieve
and von Haaren, Sebastian
and Gupta, Sagnik
and Grade, Marian
and Wirth, Matthias
and Schneider, G{\"u}nter
and Dengel, Andreas
and Ahmed, Sheraz
and Pape, Constantin},
title={Segment Anything for Microscopy},
journal={Nature Methods},
year={2025},
month={Mar},
day={01},
volume={22},
number={3},
pages={579-591},
issn={1548-7105},
doi={10.1038/s41592-024-02580-4},
url={https://doi.org/10.1038/s41592-024-02580-4}
}

@article{dinov2,
  title={Dinov2: Learning robust visual features without supervision},
  author={Oquab, Maxime and Darcet, Timoth{\'e}e and Moutakanni, Th{\'e}o and Vo, Huy and Szafraniec, Marc and Khalidov, Vasil and Fernandez, Pierre and Haziza, Daniel and Massa, Francisco and El-Nouby, Alaaeldin and others},
  journal={arXiv preprint arXiv:2304.07193},
  year={2023}
}

@misc{Bartanan,
Author = {Freida Barnatan and Emunah Goldstein and Einav Kalimian and Orchen Madar and Avi Huri and David Zitoun and Ya'akov Mandelbaum and Moshe Amitay},
Title = {Zero-shot Shape Classification of Nanoparticles in SEM Images using Vision Foundation Models},
Year = {2025},
Eprint = {arXiv:2508.03235},
}

@article{tsai2018sampler,
  title={A sampler designed for nanoparticles and respirable particles with direct analysis feature},
  author={Tsai, Candace Su-Jung and Theisen, Daniel},
  journal={Journal of Nanoparticle Research},
  volume={20},
  number={8},
  pages={209},
  year={2018},
  publisher={Springer}
}

@misc{easyocr,
  author       = {{Jaided AI}},
  title        = {EasyOCR},
  year         = {2024},
  howpublished = {\url{https://github.com/JaidedAI/EasyOCR}},
  note         = {Version 1.7.2}
}

@inproceedings{raghu2021vision,
  title={Do Vision Transformers See Like Convolutional Neural Networks?},
  author={Raghu, Maithra and Unterthiner, Thomas and Kornblith, Simon and Zhang, Chiyuan and Dosovitskiy, Alexey},
  booktitle={Advances in Neural Information Processing Systems},
  volume={34},
  pages={12116--12128},
  year={2021}
}

@inproceedings{caron2021emerging,
  title={Emerging Properties in Self-Supervised Vision Transformers},
  author={Caron, Mathilde and Touvron, Hugo and Misra, Ishan and J{\'e}gou, Herv{\'e} and Mairal, Julien and Bojanowski, Piotr and Joulin, Armand},
  booktitle={Proceedings of the IEEE/CVF International Conference on Computer Vision},
  pages={9650--9660},
  year={2021}
}

@inproceedings{dosovitskiy2020image,
  title={An Image is Worth 16x16 Words: Transformers for Image Recognition at Scale},
  author={Dosovitskiy, Alexey and Beyer, Lucas and Kolesnikov, Alexander and Weissenborn, Dirk and Zhai, Xiaohua and Unterthiner, Thomas and Dehghani, Mostafa and Minderer, Matthias and Heigold, Georg and Gelly, Sylvain and others},
  booktitle={International Conference on Learning Representations},
  year={2021}
}

@inproceedings{hariharan2015hypercolumns,
  title={Hypercolumns for Object Segmentation and Fine-Grained Localization},
  author={Hariharan, Bharath and Arbel{\'a}ez, Pablo and Girshick, Ross and Malik, Jitendra},
  booktitle={Proceedings of the IEEE Conference on Computer Vision and Pattern Recognition},
  pages={447--456},
  year={2015}
}

@article{mewada2025hybrid,
  title={Hybrid feature fusion in cervical cancer cytology: a novel dual-module approach framework for lesion detection and classification using radiomics, deep learning, and reproducibility},
  author={Mewada, H. and others},
  journal={Frontiers in Oncology},
  volume={15},
  pages={102938},
  year={2025},
  publisher={Frontiers Media SA},
  doi={10.3389/fonc.2025.102938}
}

@article{frei2023deep,
  title={Deep learning for automated size and shape analysis of nanoparticles in scanning electron microscopy},
  author={Frei, M. and Kruis, F. E.},
  journal={Beilstein Journal of Nanotechnology},
  volume={14},
  pages={58--68},
  year={2023},
  publisher={Beilstein-Institut}
}

@article{kraus2016classifying,
  title={Classifying and segmenting microscopy images with deep multiple instance learning},
  author={Kraus, O. Z. and Ba, J. L. and Frey, B. J.},
  journal={Bioinformatics},
  volume={32},
  number={12},
  pages={i52--i59},
  year={2016},
  publisher={Oxford University Press}
}

@article{adak2022imaging,
  title={Imaging atomic-scale chemistry from fused multi-modal electron microscopy},
  author={Adak, O. and others},
  journal={npj Computational Materials},
  volume={8},
  number={1},
  pages={16},
  year={2022},
  publisher={Nature Publishing Group}
}

@misc{shamshad2022transformersmedicalimagingsurvey,
      title={Transformers in Medical Imaging: A Survey}, 
      author={Fahad Shamshad and Salman Khan and Syed Waqas Zamir and Muhammad Haris Khan and Munawar Hayat and Fahad Shahbaz Khan and Huazhu Fu},
      year={2022},
      eprint={2201.09873},
      archivePrefix={arXiv},
      primaryClass={eess.IV},
      url={https://arxiv.org/abs/2201.09873}, 
}

@techreport{NIOSH_NMAM_AE_2016,
  author      = {{NIOSH}},
  title       = {NIOSH Manual of Analytical Methods (NMAM), 5th Edition, 
                 Chapter AE: Factors Affecting Aerosol Sampling},
  year        = {2016},
  institution = {National Institute for Occupational Safety and Health},
  address     = {Cincinnati, OH},
  url         = {https://www.cdc.gov/niosh/docs/2014-151/}
}

@article{Beaulieu1980OpenClosed,
  author  = {Beaulieu, H. J. and Fidino, A. V. and Arlington, K. L. B. and Buchan, R. M.},
  title   = {A comparison of aerosol sampling techniques: ``open'' versus ``closed-face'' filter cassettes},
  journal = {American Industrial Hygiene Association Journal},
  year    = {1980},
  volume  = {41},
  number  = {10},
  pages   = {758--765},
  doi     = {10.1080/15298668091425608}
}

@techreport{NIOSH_CIB65_2013,
  author       = {{National Institute for Occupational Safety and Health (NIOSH)}},
  title        = {Current Intelligence Bulletin 65: Occupational Exposure to Carbon Nanotubes and Nanofibers},
  institution  = {U.S. Department of Health and Human Services, Centers for Disease Control and Prevention, National Institute for Occupational Safety and Health},
  number       = {DHHS (NIOSH) Publication No. 2013-145},
  address      = {Cincinnati, OH, USA},
  year         = {2013},
  url          = {https://www.cdc.gov/niosh/docs/2013-145/}
}

@incollection{Birch_CNT_TEM_2017,
  author       = {Birch, M. Eileen and Wang, Chen and Fernback, Joseph E. and Feng, H. Amy and Birch, Quinn T. and Dozier, Alan K.},
  title        = {Analysis of Carbon Nanotubes and Nanofibers on Mixed Cellulose Ester Filters by Transmission Electron Microscopy},
  booktitle    = {NIOSH Manual of Analytical Methods (NMAM)},
  edition      = {5},
  publisher    = {U.S. Department of Health and Human Services, Centers for Disease Control and Prevention, National Institute for Occupational Safety and Health},
  address      = {Cincinnati, OH, USA},
  year         = {2017},
  note         = {Chapter CN}
}

@article{Oberdorster_CNTCNF_2015,
  author       = {Oberd{\"o}rster, G{\"u}nter and Castranova, Vincent and Asgharian, Bahman and Sayre, Phil},
  title        = {Inhalation exposure to carbon nanotubes (CNT) and carbon nanofibers (CNF): methodology and dosimetry},
  journal      = {Journal of Toxicology and Environmental Health, Part B: Critical Reviews},
  volume       = {18},
  number       = {3--4},
  pages        = {121--212},
  year         = {2015},
  doi          = {10.1080/10937404.2015.1051611}
}

@article{Wienke_3MAP_1995,
  author       = {Wienke, Dietmar and Xie, Yuli and Hopke, Philip K.},
  title        = {Classification of Airborne Particles by Analytical Scanning Electron-Microscopy Imaging and a Modified Kohonen Neural-Network (3MAP)},
  journal      = {Analytica Chimica Acta},
  volume       = {316},
  pages        = {239--252},
  year         = {1995},
  doi          = {10.1016/0003-2670(95)00086-L},
  note         = {Approximate page range based on available sources}
}

@article{Xie_Hopke_Wienke_1994,
  author       = {Xie, Yuli and Hopke, Philip K. and Wienke, Dietmar},
  title        = {Airborne particle classification with a combination of chemical composition and shape index utilizing an adaptive resonance artificial neural network},
  journal      = {Environmental Science and Technology},
  volume       = {28},
  number       = {11},
  pages        = {1921--1928},
  year         = {1994},
  doi          = {10.1021/es00060a024}
}

@article{Dahm_CNT_exposure_2015,
  author       = {Dahm, Matthew M. and Evans, Douglas E. and Bertke, Stephen and Birch, M. Eileen and Ku, Bon-Ki and Dunn, Kevin H.},
  title        = {Carbon Nanotube and Nanofiber Exposure Assessments: An Analysis of 14 Site Visits},
  journal      = {Annals of Occupational Hygiene},
  volume       = {59},
  number       = {6},
  pages        = {705--723},
  year         = {2015},
  doi          = {10.1093/annhyg/mev020}
}

@article{VanDerMaaten_Hinton_TSNE_2008,
  author       = {van der Maaten, Laurens and Hinton, Geoffrey},
  title        = {Visualizing data using t-SNE},
  journal      = {Journal of Machine Learning Research},
  volume       = {9},
  pages        = {2579--2605},
  year         = {2008}
}

@article{VanDerMaaten_AcceleratingTSNE_2014,
  author       = {van der Maaten, Laurens},
  title        = {Accelerating t-SNE using tree-based algorithms},
  journal      = {Journal of Machine Learning Research},
  volume       = {15},
  pages        = {3221--3245},
  year         = {2014}
}

@article{VanDerMaaten_BarnesHutSNE_2013,
  author       = {van der Maaten, Laurens},
  title        = {Barnes-Hut-SNE},
  journal      = {arXiv preprint arXiv:1301.3342},
  year         = {2013}
}

@article{van2014scikit,
  title={scikit-image: image processing in {P}ython},
  author={van der Walt, St{\'e}fan and Sch{\"o}nberger, Johannes L and
          Nunez-Iglesias, Juan and Boulogne, Fran{\c{c}}ois and
          Warner, Joshua D and Yager, Neil and Gouillart, Emmanuelle and
          Yu, Tony and the scikit-image contributors},
  journal={PeerJ},
  volume={2},
  pages={e453},
  year={2014},
  publisher={PeerJ Inc.},
  doi={10.7717/peerj.453}
}

@ARTICLE{Dahm2018-vh,
  title    = "Exposure assessments for a cross-sectional epidemiologic study of
              {US} carbon nanotube and nanofiber workers",
  author   = "Dahm, Matthew M and Schubauer-Berigan, Mary K and Evans, Douglas
              E and Birch, M Eileen and Bertke, Stephen and Beard, John D and
              Erdely, Aaron and Fernback, Joseph E and Mercer, Robert R and
              Grinshpun, Sergey A",
  journal  = "Int. J. Hyg. Environ. Health",
  volume   =  221,
  number   =  3,
  pages    = "429--440",
  month    =  apr,
  year     =  2018,
  language = "en"
}

@techreport{niosh7400,
  author       = {Baron, Paul A.},
  title        = {{Asbestos and Other Fibers by PCM: Method 7400}},
  institution  = {National Institute for Occupational Safety and Health (NIOSH)},
  year         = {1994},
  month        = aug,
  type         = {Issue 2},
  number       = {Method 7400},
  note         = {In NIOSH Manual of Analytical Methods (NMAM), 4th Edition},
  url          = {https://www.cdc.gov/niosh/docs/2003-154/pdfs/7400.pdf},
}
\bibliographystyle{rsc}

\newpage


\end{document}